\documentclass[twoside,11pt]{article}

\usepackage{jair, theapa, rawfonts}
\usepackage{framed}
\usepackage{url}
\usepackage{graphicx}

\begin{document}

\title{Emergent Multi-Agent Communication in the Deep Learning Era}

\author{\name Angeliki Lazaridou \email angeliki@google.com \\
       \addr DeepMind \\6 Pancras Square, London, UK       
       \AND
       \name Marco Baroni \email mbaroni@gmail.com \\
       \addr Facebook AI Research, Rue M\'enars 6, Paris, France \\
       \addr ICREA, Passeig Llu\'is Companys 23, Barcelona, Spain \\
       \addr Departament de Traducci\'o i Ci\`encies del Llenguatge, UPF, Roc Boronat 138, Barcelona, Spain
       }

\maketitle

\begin{abstract}
  The ability to cooperate through language is a defining feature of humans. As
the perceptual, motory and planning capabilities of deep artificial networks
increase, researchers are studying whether they also can develop a shared
language to interact. 
From a scientific perspective, understanding the conditions under which
language evolves in communities of deep agents and its emergent features can shed light on human language evolution. From an applied perspective, endowing deep networks
with the ability to solve problems interactively by communicating with each
other and with us should make them more flexible and useful in everyday life. 
 This article surveys representative recent language emergence studies from
 both of these two angles.

\end{abstract}

 
\begin{framed}
  \section*{Highlights}
  \begin{itemize}
  \item Deep networks and techniques from deep reinforcement learning
    have greatly widened the scope of computational simulations of language
    emergence in communities of interactive agents.
  \item Thanks to these modern tools, language emergence can now be
    studied among agents that receive realistic perceptual input, must
    solve complex tasks cooperatively or competitively, and can engage
    in flexible multi-turn verbal and non-verbal interactions.
  \item With great simulation power comes great need for new analysis
    methods: a budding area of research focuses on understanding
    the general characteristics of the deep agents' emergent language.
  \item Another line of research wants to deliver on the promise of
    interactive AI, exploring the functional role of emergent
    language in improving machine-machine and human-machine communication.
  \end{itemize}
\end{framed}

\section{Introduction}

The last decade has seen astounding progress in the development of
artificial neural networks, under the ``deep learning'' rebranding
\cite{LeCun:etal:2015}.  In computer vision, we can now
automatically recognize thousands of objects in natural images
\cite{Russakovsky:etal:2014,Krizhevsky:etal:2017}. In the domain of
natural language, deep networks led to great progress in applications
ranging from machine translation to document understanding
\cite{Vaswani:etal:2017,Edunov:etal:2018,Devlin:etal:2019,Radford:etal:2019}. %
Deep networks that combine vision and language can generate image
captions and answer complex questions about scenes with high accuracy
\cite{Anderson:etal:2018,Zhou:etal:2020}.

These successes are attained by networks that are passively exposed to massive
amounts of text and/or images, and learn to rely on the
statistical regularities they extracted from their training data. %
The interactive, functional aspects of language and intelligence \cite<e.g.,>{Wittgenstein:1953,Austin:1962,Searle:1969,Clark:1996,Allwood:1976,Pickering:Garrod:2004,Linell:2009,Ginzburg:Poesio:2016}
are completely ignored. By definition, an \emph{agent} cannot
learn to (inter-)\emph{act} just by being passively exposed to
lots of data (even when these data are records of
interactions). An exclusive focus on passive statistical learning has practical consequences. Despite much
exciting work in the area, deep-learning-based chatbots and dialogue
systems \cite{Serban:etal:2016,Gao:etal:2019} are still extremely limited in their
capabilities, missing on the dynamic, interactive nature of conversation
\cite{Bernardi:etal:2015}. And AI agents able to fully cooperate with humans
are still a science-fiction
dream \cite{Mikolov:etal:2016}.

\begin{figure}[p]
  \centering
\includegraphics[width=14cm]{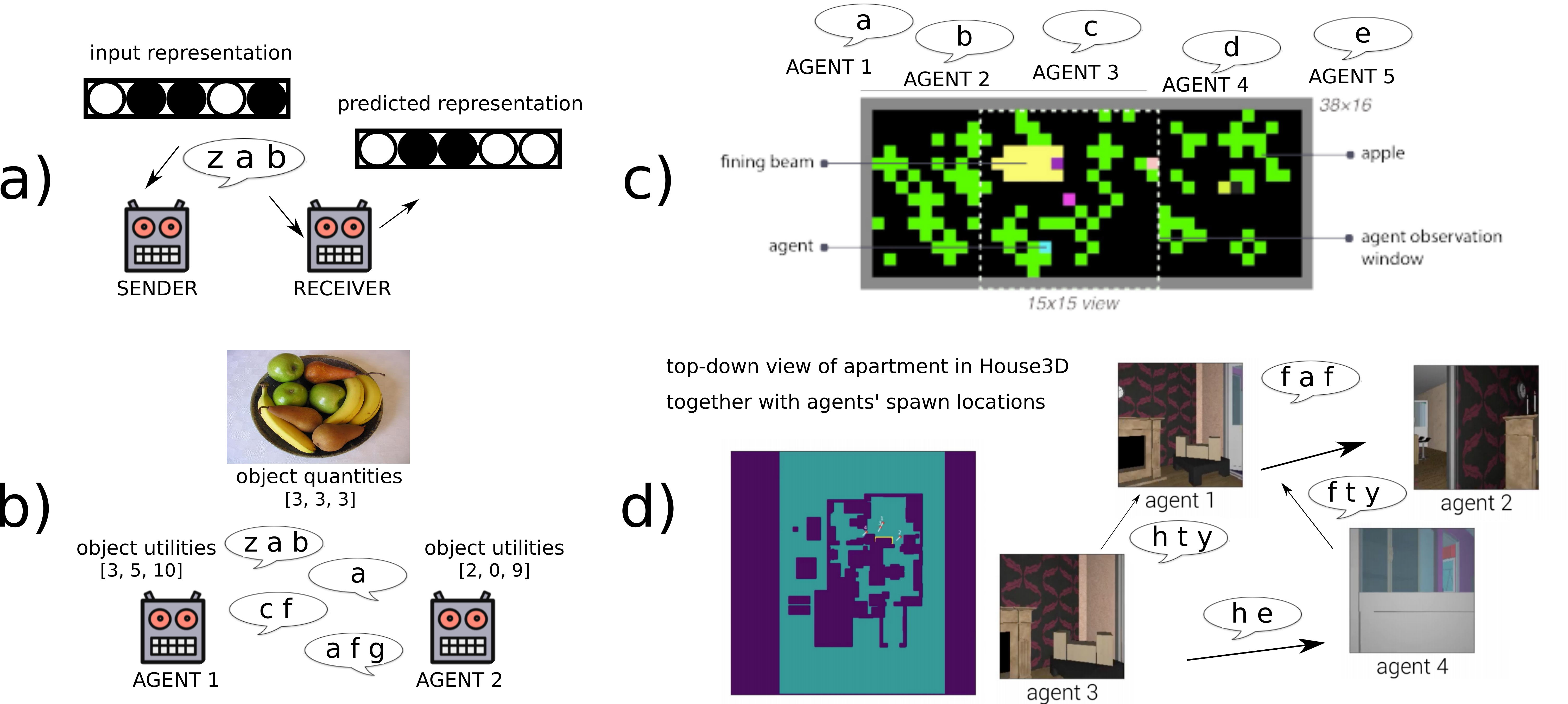}
\caption{\textbf{Examples of games and environments for emergent communication.}
(a) Emergent communication work in the pre-deep-learning era typically used symbolic data as input: \citeA{Batali:1998} presents a study where recurrent neural network agents communicate in a referential game using sequences of discrete symbols. Similar work with deep networks often uses realistic pictures as input, see Fig.~\ref{fig:referential-game} for an example. (b) More complex scenarios with deep agents: \citeA{Cao:etal:2018} study self-interested agents engaging in a multi-turn negotiation game. (c) Richer, dynamic environments: \citeA{Jaques:etal:2019} study five embodied self-interested agents engaging in multi-turn interactions while navigating in a 2D visual environment. (d) Scaling up to fully realistic scenarios: in the experiment of \citeA{Das:etal:2019}, embodied cooperative agents solve navigation challenges in a 3D environment. Images from \citeA{Jaques:etal:2019} and \citeA{Das:etal:2019} reproduced by permission.}\label{fig:environments}
\end{figure}

The aim of developing devices capable of genuine linguistic interaction has revived interest in
studying the ``languages'' actively
developed by communities of artificial agents that must communicate in
order to succeed in their environment. Earlier
work in this mold \cite<e.g.,>{Cangelosi:Parisi:2002,Christiansen:Kirby:2003,Steels:2003b,Wagner:etal:2003,Steels:2012,Hurford:2014} 
explored very specific questions through carefully designed experimental
simulations that involved simple, largely hand-crafted agents. For example, \citeA{Batali:1998}  performed simulations in which a sender agent, given an input binary vector representing the meaning of a simple phrase (e.g., \textit{you smile}), encodes it as a sequence of characters. These are transmitted to a receiver agent, who needs to decode the original meaning (Fig.~\ref{fig:environments}(a)). The question under investigation was whether agent messages would mirror the grammatical structure encoded in the simple input phrase, as in natural language, and this turned out indeed to be the case.

\begin{figure}[p]
  \centering
  \includegraphics[width=14cm]{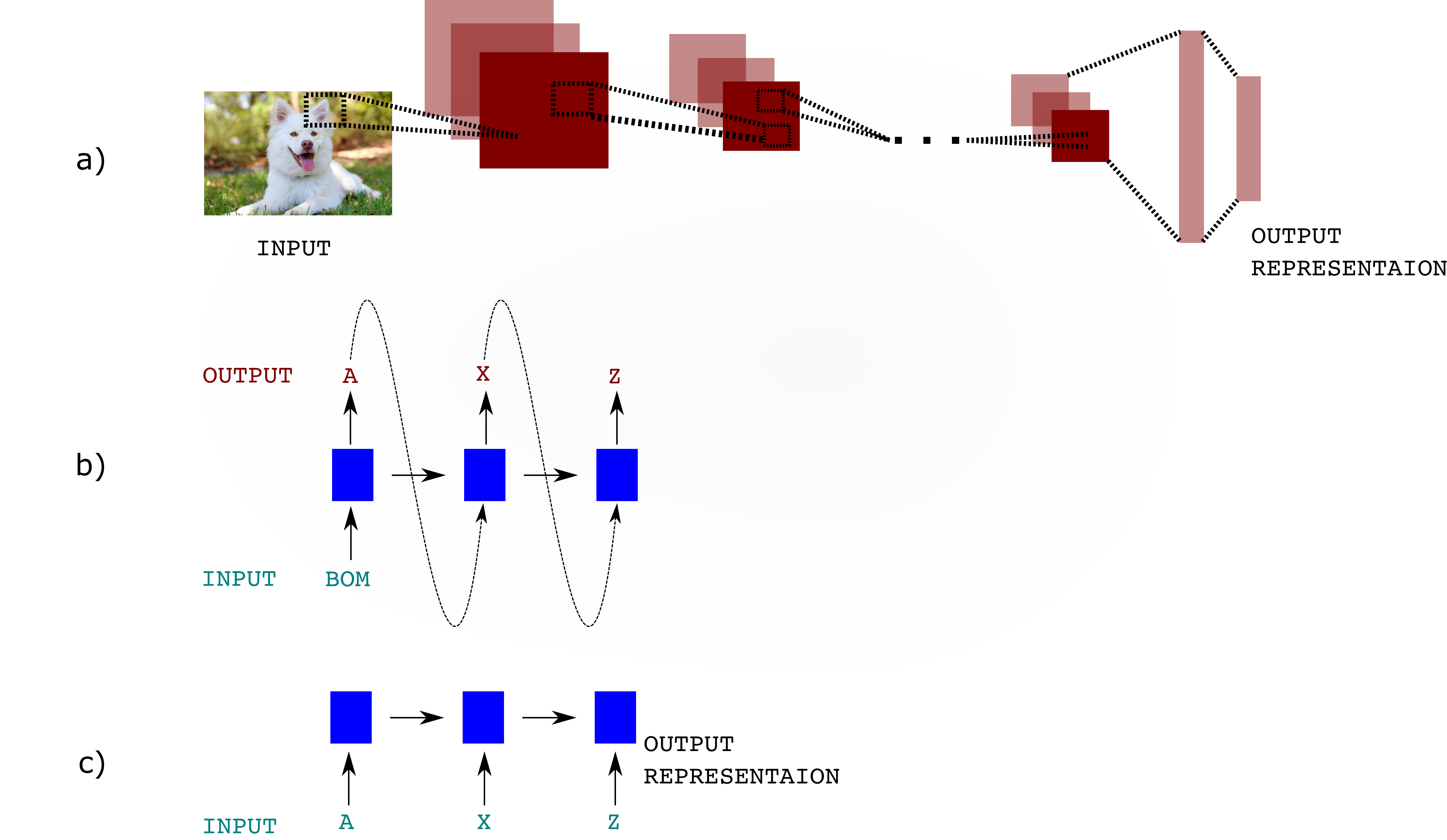}
  \caption{\textbf{Typical neural network components of a deep agent.}
    (a) A \emph{visual processing} module (typically a convolutional
    network) converting pictures into internal distributed
    representations.  (b) A \textit{generation} component consisting
    of a recurrent neural network that produces a symbol sequence
    (in this case, $AXZ$).  (c) An
    \textit{understanding} module, that takes as input a sequence of
    units (in this case, the symbols produced by the generation
    component) and produces an internal distributed representation.  A typical
    \textit{sender} agent will first transform images into distributed
    representations with (a) and then use (b) to produce a message.
    A \textit{receiver} agent will also use (a) to transform images to
    representations, and then (c) to process the message from the
    sender in order to make a decision about the output action. In
    both cases, further layers are interspersed with the
    various components to further aid the agents' ``reasoning''
    process (e.g., the receiver might use them to combine visual and verbal information).
    \label{fig:agents}}
\end{figure}

\begin{figure}[p]
  \centering
\includegraphics[width=10cm]{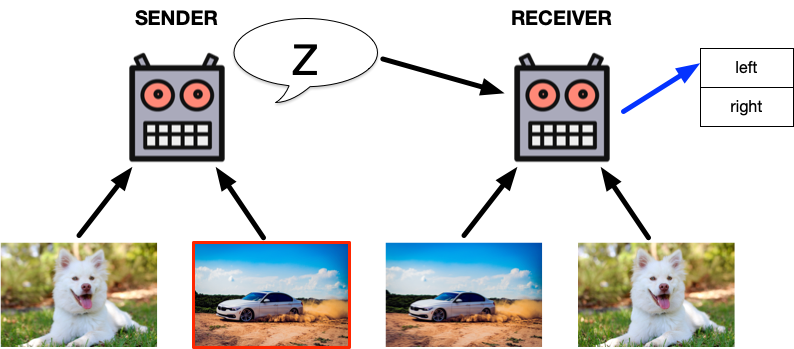}
  \caption{\textbf{The referential game of  \citeA{Lazaridou:etal:2017}.} In a referential game, successful communication is the very purpose of the game (as opposed to scenarios in which communication can help players to achieve an independent goal, such as obtaining a valuable object). Referential games have a long history in linguistics, philosophy and game theory \cite{Lewis:1969,Skyrms:2010}. In the game illustrated here, the sender network receives in input two natural images, depicting instances of two distinct categories out of about 500 (here: a dog and a car), with one of the images marked as target (here, the car). The sender processes the images with a convolutional network module and it emits one symbol (sampled from a fixed alphabet), that is given as input to the receiver network, together with the two images (in random order). If the receiver ``points'' to the correct location of the target in the image array (as it does in the figure), both agents are rewarded. The networks are trained by letting them play the game many times, and adjusting their weights based on the reward signal. No supervision is provided about the symbols to be used for communication, so that they are completely free to adapt the emergent protocol to their strategies and biases.}
  \label{fig:referential-game}
\end{figure}

Today,
generic ``deep agents'' (Fig.~\ref{fig:agents}), built out of standard
components such as convolutional and recurrent
  networks \cite{LeCun:etal:1998,Elman:1990,Hochreiter:Schmidhuber:1997}, with little or no task-specific tweaking,
are being used in simulations that go beyond what was conceivable just
a few years ago: dealing with complex scenarios involving thousands of
possible referents, that are presented in perceptually realistic
formats; engaging in self-paced multi-turn interactions; producing
long, language-like utterances (Fig.~\ref{fig:environments},
Fig.~\ref{fig:referential-game}).


In this survey, we review this recent literature on language emergence
in deep agent communities. After introducing some representative examples, we focus specifically on two lines of
investigation that are currently prominent in the field. First, the very complexity and richness of deep agents and their
environments implies that they often will succeed at communicating while using very opaque codes. This has led
to extensive analytical work on decoding the emergent protocol, in an attempt to understand its generality and its similarities to human language (if any), and to identify possible degenerate cases.

Sceond, we review studies that focus on how to make emergent language more powerful and
useful from an AI perspective. This involves, on the one hand, exploring
how a self-induced communication protocol might benefit deep networks
endowed with advanced perceptual and navigational capabilities. On the
other, researchers are studying how to let agents evolve  more
human-like languages, with the aim of establishing effective human-machine
communication.

\section{Language Emergence in Deep Agent Communities}
\label{sec:new-simulations}

In the representative simulations we will describe in this section, the agents are presented with a
 task, and each agent has a cost or reward function to optimize.
Agents have perfectly aligned incentives, i.e., they
 share their reward. Communication comes into play as a means to
achieve their goal. Learning generally takes place through %
reinforcement learning, a set of techniques to train systems in scenarios in which the main teaching signal is \emph{reward} for succeeding or failing at a task, possibly requiring multiple actions in a potentially changing environment \cite{Sutton:Barto:1998,Mnih:etal:2015,Silver:etal:2016}. This setup offers more flexibility than standard \emph{supervised} learning (where the learning signal derives from direct comparison of the system output with the ground-truth), but it is also more challenging. Communication is emergent in the sense that, at the beginning of a simulation, the
 symbols the agents emit have no \emph{ex-ante} semantics nor pre-specified usage rules. Meaning and syntax emerge through game play. %

\subsection{Continuous and Discrete Communication}
\label{sec:discrete-continuous}
 Communication can be of two types:
  \textit{continuous}, in which agents communicate via a continuous
  vector, and \textit{discrete}, in which agents
  communicate by means of single symbols or sequences of symbols.
An example of the continuous case is the influential DIAL system of \citeA{Foerster:etal:2016}. The agents are given a continuous communication channel, making it easy to back-propagate learning signals through the whole system. A continuous vector connecting two agent networks can equivalently be seen as another activation layer in a larger
architecture encompassing the two networks, and it effectively gives each agent access to the internal states of the other network. Therefore, continuous communication turns the multi-agent system into a single large network.  
A ``vanilla'' model of discrete communication commonly used in the language emergence literature and for multi-agent coordination problems is RIAL \cite{Foerster:etal:2016}. In RIAL, communication happens through discrete symbols, thus making it impossible for agents to transmit rich error information via continuous back-propagation through each other. The only learning signal received by each agent is task reward. As such, unlike in the continuous case, and similarly to what happens in human communities, each agent treats the other(s) as part of its environment, with no access to their internal states. It is thus exactly the presence of the discrete bottleneck that makes simulations genuinely ``multi-agent''.  Moreover, communication via discrete symbols provides the symbolic scaffolding for interfacing the agents' emergent code to natural language,  which is universally discrete \cite{Hockett:1960}.
Tuning the weights of a neural network with an error signal that is back-propagated through a discrete bottleneck is a challenging technical problem. Adopting methods from reinforcement learning, agent training can take place using the REINFORCE update rule \cite{Williams:1992,Lazaridou:etal:2017}, which intuitively increases the weight for actions that resulted in a positive  reward  (proportional to their probability), and decreases them otherwise. Alternatively, discrete representations can be approximated by continuous ones during the training  phase \cite{Havrylov:Titov:2017,Jang:etal:2017,Maddison:etal:2017}.

\subsection{Representative Studies}

Since we initially focus on the comparison of
emergent codes with natural language,
we briefly review here work that considers the discrete case, coming back
to some examples of continuous communication in Section
\ref{sec:agent-coordination} below.


One of the first studies of language emergence in deep networks was presented
by \citeA{Lazaridou:etal:2017}, who used the referential game schematically
illustrated in Fig.~\ref{fig:referential-game}. %
%
This paper first
showed that agents can develop an effective communication protocol to talk about realistic
images by relying on game success as sole training signal. Still, evidence that the agents were developing human-like words referring to generic concepts such as ``dog'' or ``animal'' was mixed,
a point will return to in the next section.

While \citeA{Lazaridou:etal:2017} constrained messages to consist of one
symbol, \citeA{Havrylov:Titov:2017} allowed the sender to emit strings
of symbols of variable length
\cite<see also>{Lazaridou:etal:2018}. The resulting emergent language developed
a prefix-based hierarchical scheme to encode meaning into
multiple-symbol sequences. For example, the ``word'' for pizza was
\emph{5261 2250 5211}, where \emph{5261} refers to food, \emph{2250}
to baked food, and \emph{5211} to pizzas. 

\citeA{Evtimova:etal:2018} went one step further, considering multiple-turn
interactions \cite<see also>{Jorge:etal:2016}.
In their game, one agent
must pick the definition of an animal from a list of dictionary
entries, when a natural image of the target animal is presented to
the other agent. 
The agents can exchange
multiple messages. The conversation ends when
the agent tasked with guessing the definition makes its
final guess.  Several natural
properties of conversations emerged in this setup. For example, the
agents tend to exchange more turns in more difficult game
episodes. 

A further step towards realistic conversational scenarios, beyond
referential games, is taken by \citeA{Bouchacourt:Baroni:2019} \cite<see also>{Cao:etal:2018}. In this study, one agent is assigned a
fruit, the other two tools, and their task is to decide which of the
two tools is best for the current fruit. The utility of each tool with
respect to each fruit is derived from a corpus of human judgments,
resulting in skewed affordance statistics (e.g., a knife is generally
more useful than a spoon). The setup is fully symmetric, with either
agent randomly assigned either role in each episode, and both agents
being able to start and end the conversation. The agents learn to use
messages meaningfully, accumulating more reward than what they could
get by relying on general object affordances. However, despite the
symmetric setup, they develop different idiolects for the different
roles they take, that is, the same agents use different codes to
communicate the same meanings, depending on who is in charge of describing the fruit, and who the tools. A similar behaviour  was also observed by \citeA{Cao:etal:2018}.

Under which conditions will agents converge to a shared language is
one of the topics addressed by \citeA{Graesser:etal:2019}, who used deep agent communities as a
modeling tool for contact linguistics
\cite{MyersScotton:2002}. They report that,
if two agents will develop different idiolects, it suffices for
the community to include a third agent for a shared code to
emerge. One of their most interesting findings is that, when agent
communities of similar size are put in contact, the agents develop a
mixed code that is simpler than either original language, akin
to the development of pidgins and creoles in mixed-language
communities \cite{Bakker:etal:2011}.

\section{Understanding the Emergent Language}
\label{sec:understanding-emergent-language}

With more realistic simulations, understanding
what is going on becomes more difficult. Even when we are confident
that genuine communication is taking place (Section
\ref{sec:measuring-communication}), it is difficult to decode messages
by simple inspection. We do not know if and how the messages produced
by the agents should be segmented into ``words''. We might only have
vague conjectures about what they refer to. If there are multiple
turns, we do not know which turns are mostly information exchanges,
and which, if any, absolve other pragmatic functions (e.g., asking for more information). We cannot even
trust the agents to use symbols in a consistent way across contexts
and turns \cite{Bogin:etal:2018}. The enterprise is akin to
linguistic fieldwork, except that we are dealing with an alien race,
with no guarantees that universals of human communication will
apply.

\begin{figure}[p]
  \centering
\includegraphics[width=7cm]{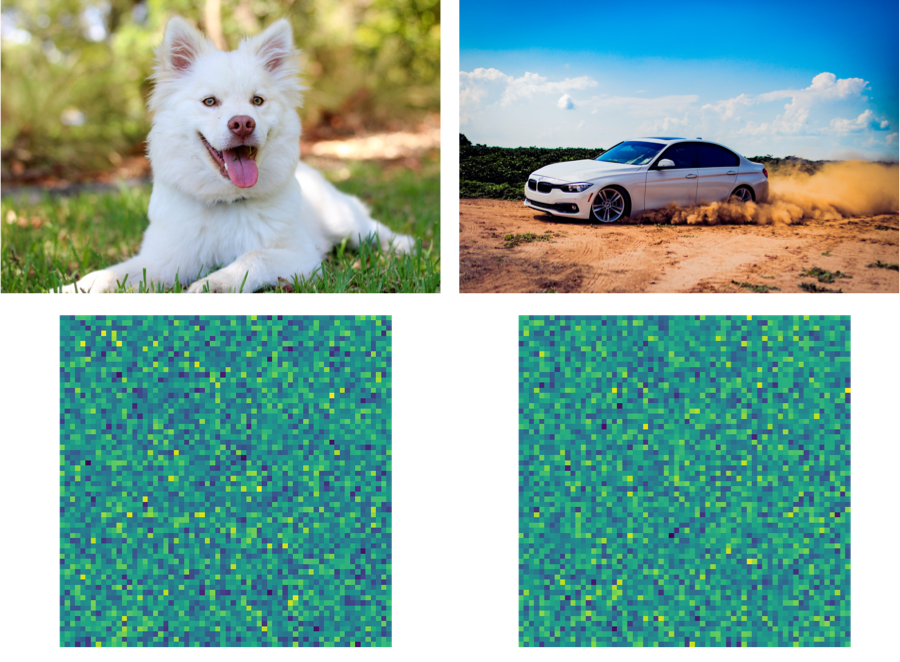}
\caption{\textbf{Training and test inputs in the referential game of
    \cite{Bouchacourt:Baroni:2018}.} Two agents were trained to play
  the  game of \citeA{Lazaridou:etal:2017} (see
  Fig.~\ref{fig:referential-game}). During training, the agents were
  exposed to the same data as in the original study, that is, pairs of
  pictures of instances of about 500 distinct objects (top row). At
  test time, however, the agents were made to play the game with blobs
  of Gaussian noise (bottom row). They were able to
  communicate about them nearly as well as about the training
  pictures. This shows that the language emerging in this game does
  not involve ``words'' referring to generic concepts, but rather
  \emph{ad-hoc} signals, probably carrying comparative information
  about shallow visual properties of the images. Bottom row reproduced
  from \citeA{Bouchacourt:Baroni:2018} by permission.}
  \label{fig:gaussian}
\end{figure}

Indeed, in spite of task success, the emerging language can have
counter-intuitive properties. \citeA{Kottur:etal:2017} considered
agents playing a referential game in which they must communicate about
object attributes and values (e.g, \emph{color: blue}, \emph{shape:
  round}). The agents have difficulties converging to the intuitive
coding scheme in which distinct symbols unambiguously denote single
attributes or values (i.e., a word for \emph{color}, a word for
\emph{blue}, etc). Such code will only emerge when the set of
available symbols is greatly limited and the memory of one of the
agents is ablated, pointing to memory bottlenecks as a possible
bias to be injected into deep networks for more natural languages to
emerge \cite<see also>{Resnick:etal:2020}. \citeA{Bouchacourt:Baroni:2018} replicated
the game of \citeA{Lazaridou:etal:2017} with the surprising results
illustrated in Fig.~\ref{fig:gaussian}.

Essentially, agents will develop a code that is sufficient to solve
the task at hand, and hoping that such code will possess further
desirable characteristics is wishful thinking. Consider the task in
the original game of \citeA{Lazaridou:etal:2017}. The agents must
discriminate pairs of pictures depicting instances of 500
categories. The agents could achieve this by developing human-like
names for the categories, but a low-level strategy relying on, say,
comparing average pixel intensity in patches of the two images might
require as few symbols as 2. In this respect, the
agents' language is, paradoxically, ``too human'', in the sense that
it evolved to minimize effort, while remaining adequate for the task
at hand \cite{Gibson:etal:2019}. 
Indeed, \citeA{Kharitonov:etal:2020} showed
that the way deep agent emergent languages partition their
meaning space displays the same tendency towards complexity minimization
that is pervasive in human language.

\citeA{Chaabouni:etal:2019} studied whether agent language
exhibit an inverse correlation between word frequency and word length,
so that the signals that need to be used more often are also the
shortest, as universally found in natural languages
\cite{Zipf:1949,Strauss:etal:2007,FerrerICancho:etal:2013}. They discovered that deep agents trained with a referential game where inputs
have a skewed distribution similar to natural language actually
develop a significantly \emph{anti-efficient} code, in which the
most frequent inputs are associated to the longest messages. The
effect is explained by the lack of an articulatory effort
minimization bias in networks, that are thus only subject to a
``perceptual'' pressure favoring longer messages, as they are easier to discriminate.
\subsection{Measuring the Degree of Effective Communication}
\label{sec:measuring-communication}
  As simulations move beyond referential games (where task success
  trivially depends on establishing a communication code), to complex
  environments where communication plays an auxiliary function (e.g.,
  \citeA{Das:etal:2019}; Fig.~\ref{fig:environments}(d)), the first
   question to ask when analyzing an emergent language is whether
  it is actually been used in any meaningful way by the agents.  As
  clearly discussed by \citeA{Lowe:etal:2019}, just ablating the
  language channel and showing a drop in task success does not prove
  much, as the extra capacity afforded by the channel architecture
  might have helped the agents' learning process without being used to
  establish communication. The same paper proposes a classification of
  measures to detect the presence of genuine
  communication. \textit{Positive signaling} captures the extent to
  which information about the sender states, observations and actions
  are expressed in its signals. \textit{Positive listening} captures
  the extent to which a signal impacts the receiver's states and
  behaviour. Examples of positive signaling include \textit{context
    independence}~\cite{Bogin:etal:2018} and \textit{speaker
    consistency}~\cite{Jaques:etal:2019}. The former measures the
  degree of alignment between messages and task-related concepts,
  whereas the latter measures, through mutual information, the
  alignment between an agent messages and its actions. Positive
  signaling gives no guarantee of communication, since the receiver could
  be ignoring the sender messages, no matter how informative they might
  be. An example of positive listening is the \textit{instantaneous
    coordination} measure of~\citeA{Jaques:etal:2019}, which uses
  mutual information to quantify correlation between 
  sender messages and receiver actions. Instead,~\citeA{Lowe:etal:2019}
  propose to use \emph{causal influence of communication}, a
  quantity that measures the causal relationship between sender
  messages and receiver actions.   The authors show that only a high causal influence
  of communication is both necessary and sufficient for
  positive listening, and thus communication.

  The call for caution
  is not just hypothetical. Several studies have reported how agents
  can easily converge to non-verbal or degenerate strategies, even
  when it would seem that communication is taking place. For example,
   agents might learn to exchange information simply through the
  number of turns they take before ending the game, irrespective of
  what they actually say
  \cite{Cao:etal:2018,Bouchacourt:Baroni:2019}.

\subsection{Compositionality}


Much analytical work in the area has focused on compositionality, as the latter is seen both as a
fundamental feature of natural language whose evolutionary origins are
unclear \cite{Bickerton:2014,Townsend:etal:2018}, and as a pre-condition
for an emergent language to generalize at scale.

The simplest way to probe for compositionality in an emergent protocol
is to test whether agents can use it to denote novel composite
meanings, e.g., can they refer to \emph{blue squares} on first encounter, if
they have seen other \emph{blue} and \emph{square} things during
training \cite{Choi:etal:2018}. This assumes that a
compositional encoding is necessary to generalize. However, a few recent papers have intriguingly reported %
that emergent languages can support generalization to novel composite meanings \emph{without} conforming to even weak notions of
compositionality \cite{Lazaridou:etal:2018,Andreas:2019,Chaabouni:etal:2020}. 
\citeA{Lazaridou:etal:2018} re-introduced to the emergent language
community the \emph{topographic similarity} score from earlier work on
language emergence \cite{Brighton:2006}. Given ways to measure
distances between meanings and between forms, topographic similarity
is the correlation between all possible meaning pair distances and the
distances of the corresponding message pairs. It captures the intuition that compositionality involves a systematic
relation between form and meaning similarity. However, it does not tell us
anything about the nature of the specific compositional processes present
in a language. %
\citeA{Andreas:2019}
proposed a method to quantify to what extent an emergent language reflects specific types of compositional structure in its
input. Unfortunately, the method only works if we have a concrete
hypothesis about the underlying composition function, that is, it can only be used to test whether a language conforms to an underlying compositional grammar if we are able to precisely specify this grammar. This limits its practical
applicability. Finally, \citeA{Chaabouni:etal:2020} recently established a link between compositionality and the notion of disentanglement in representation learning \cite{Suter:etal:2019}, and proposed to use methods to quantify disentanglement from that literature in order to measure the degree of compositionality of emergent codes.

Equipped with similar tools, various studies have uncovered different aspects
of compositionality in emergent languages. For example, \citeA{Lazaridou:etal:2018}
found that compositionality more easily emerges when objects are
represented symbolically as sets of attribute-value pairs, than when
they are more realistically represented as synthetic 3D shapes. %
\citeA{Mordatch:Abbeel:2018}  studied
the code emerging in a community of agents moving and
acting in a shared grid-world. Each agent was assigned a goal, that
could involve having another agent moving to a landmark position. An
order-insensitive concatenative language emerged, where agents would
refer to actions, their agent and targets by juxtaposing specialized
symbols (e.g., one message could be: \emph{goto blue-agent
  red-landmark}, or, equivalently, \emph{red-landmark goto
  blue-agent}). 

Despite many intriguing empirical observations, our characterization
of which architectural biases and environmental pressures favour the
emergence of compositionality (or other linguistic properties) is
still very sketchy. A strong result obtained both with humans and in
pre-deep-learning computational simulations is that generational
transmission of language favors compositionality
\cite<e.g.,>{Kirby:Hurford:2002,Kirby:etal:2014}, an observation recently
confirmed for deep agents \cite{Li:Bowling:2019,Ren:etal:2019}.
Moreover, recent results from experiments with humans demonstrate that
larger communities of speakers evolve more systematic languages
\cite{Raviv:etal:2019a,Raviv:etal:2019b}, suggesting the need to
move away from two-partner agent setups, an observation
beginning to find its way into deep agent research
\cite{Graesser:etal:2019,Tieleman:etal:2019}.

Other priors for compositionality that have been proposed and at least partially
empirically validated include input representations
\cite{Lazaridou:etal:2018}, agent and channel capacity
\cite{Kottur:etal:2017,Mordatch:Abbeel:2018,Resnick:etal:2020}, and
specific training strategies, such as letting the agents simulate
other agents' understanding of one's language
\cite{Choi:etal:2018}. We still lack, however, systematic
experiments establishing which of these conditions are necessary,
which are sufficient, and how they interact, possibly along the lines
of earlier systematizing work such as \citeA{Bratman:etal:2010}.

\section{Emergent Communication for Better AI}
\label{sec:frontiers}

 
\subsection{Communication Facilitating Inter-Agent Coordination}
\label{sec:agent-coordination}

A number of sequential decision-making problems in communication networks \cite{Cortes:etal:2004}, finance \cite{Lux:Marchesi:1999} and other fields cannot be tackled without multi-agent modeling. As the complexity of tasks and the number of agents grow, the coordination abilities of agents become of fundamental importance. Humans excel at large-group coordination, and language clearly plays a central role in their problem solving ability \cite{Tomasello:2010,DavidBarrett:Dunbar:2016,Lupyan:Bergen:2016}. This insight is inspiring algorithmic innovations in multi-agent learning, where communication is used to facilitate coordination among multiple agents interacting in complex environments. While many ingredients of the experiments we review in this section are shared with those we discussed above, now our emphasis is not on the nature of the emergent language, but on whether its presence will aid multi-agent communities to achieve better coordination. Consequently, we focus on setups going beyond referential games, looking at what communication brings in terms of ``added value'' when it is not simply a goal in itself.

Pre-deep-learning work on multi-agent communication for coordination
\cite{Panait:Luke:2005} used to hard-code communication, e.g., by directly
sharing sensory observations or information concerning the current
state of agents or their policies
\cite{Tan:1993}. 
Reinforcement learning provides a mechanism for learning communication
protocols, lifting many assumptions required by hand-coded
protocols \cite{Kasai:etal:2008}. \citeA{Foerster:etal:2016}  combined reinforcement
learning and deep networks in the context of developing communication
protocols for interacting agents, presenting experiments with both
discrete and continuous communication (the RIAL and DIAL systems briefly discussed in Section \ref{sec:discrete-continuous} above). The study found that allowing agents to communicate improves coordination, as indicated by
higher team rewards compared to no-communication controls. However,
while continuous communication systematically results in improved
coordination \cite<see also>{Sukhbaatar:etal:2016,Kim:etal:2019,Singh:etal:2019},
discrete communication does not yield consistent improvements when the
complexity of the environment grows, and it only manages to marginally improve
on the baselines when the agents are constrained to share the same
weight parameters, a rather unrealistic assumption.

Learning with a discrete channel is more challenging due to the joint exploration problem, i.e., the environment non-stationarity introduced by the fact that all agents are learning  simultaneously and independently.  In an attempt to facilitate learning with a discrete channel, \citeA{Lowe:etal:2017}  allowed centralized training but decentralized execution. Specifically, the authors modified the standard actor-critic approach from reinforcement learning \cite{Sutton:Barto:1998}, under which an agent's own observation and action are used by an agent-specific ``critic'' to produce an estimate of the value of the action. In \citeA{Lowe:etal:2017}, the critic was shared by all agents,  thus allowing them to receive, at training time only, extra information about the other agents' policies, without access to their internal states.  

In the already discussed study of \citeA{Mordatch:Abbeel:2018}, agents are placed in an environment in
which they can use non-verbal means of communication, i.e., communicate
directly through their actions, much like the bees' waggle dance
\cite{VonFrisch:1967}. When explicit verbal
communication is disallowed, the agents find other means to
coordinate, such as pointing, guiding and pushing. While more
restrictive than a proper language relying on its own separate channel, this
type of communication might be easier to learn, as actions are already
grounded in the agents' environment, unlike linguistic communication,
which assumes utterances to carry no \emph{ex-ante} semantics. It is a natural question whether non-verbal communication could act as a
stepping stone towards more complex forms of language.

Finally, the importance of looking at human communication as source of inspiration for inductive biases has not gone unnoticed. \citeA{Eccles:etal:2019} capitalize on pragmatics, considering a speaker whose goal is to be informative and relevant (adhering to the equivalent Gricean maxims), and a listener who assumes that the speaker is cooperative, i.e., providing meaningful and relevant information \cite{Grice:1975}.  The authors frame these inductive biases into additional training objectives, one for each interlocutor. The speaker is rewarded for message policies that have high mutual information with the speaker's trajectory,  resulting in the production of different messages in different situations. The listener is rewarded when its behaviour is affected by the speaker's messages, encouraging it to attend to the communication channel. \citeA{Foerster:etal:2019} simulate a process akin to pragmatic inference \cite{Goodman:Frank:2016}, which guides human behaviour in a variety of communicative scenarios. Specifically,  in the context of Hanabi \cite{Bard:etal:2020}, a cooperative card game where agents communicate through their game moves, agents reason about their co-players' observable actions, aiming at uncovering their intents and modeling their beliefs, in order to produce more informative signals.

\subsection{Beyond Cooperation: Self-interested and Competing Agents}

While most deep agent emergent communication work considers
interactions between cooperative agents, there is increasing interest in cases where
agents' interests diverge. Communication between self-interested and
competing agents has been extensively studied in game theory and
behavioral economics \cite{Crawford:Sobel:1982}, since in human
interaction and decision making tensions between collective and
individual rewards constantly arise. From a practical point of view, a
better understanding of emergent communication in
non-fully-cooperative situations can positively impact applications
such as self-driving cars.

Theoretical results suggest that,
when the agents' incentives are not aligned, meaningful communication
is not guaranteed \cite{Farrell:Rabin:1996}. Compared to what happens when agents
communicate directly through their core actions \cite<as in>{Mordatch:Abbeel:2018}, linguistic communication differs in
three key properties,  labeled  as ``cheap talk'' by
\citeA{Farrell:Rabin:1996}. Linguistic communication
is i) costless, i.e., the sender incurs no penalty for sending
messages; ii) non-binding, i.e., messages sent through this channel do
not commit the sender to any course of action and iii) non-verifiable,
i.e., there is no inherent link between linguistic communication and
the agents' behaviours, so that agents can potentially lie. %
In cooperative games this is not an issue, since the agents' incentives are aligned and communication can only increase their pay-offs. When agent interests diverge, however, senders could choose to communicate information increasing their personal reward only (and potentially decreasing that of others), and consequently disincentivize receivers from paying attention.

\citeA{Cao:etal:2018} study language emergence in a
semi-cooperative model of agent interaction, i.e., a negotiation
environment \cite{DeVault:etal:2015} consisting in a multi-turn
version of the ultimatum game \cite{Guth:etal:1982}. In each episode,
agents are presented with a set of objects, and each agent is
assigned a hidden value for each object (e.g., in an episode, peppers
might be very valuable for one agent, and cherries useless).  At each
step, agents emit a cheap talk message, as well as a (non-verbally
conveyed) proposal on how to split the goods. Either agent can
terminate the episode at any time by accepting the proposal that the
other agent made in the previous step. If the agents do not reach an
agreement within 10 turns, neither gets any reward. The authors find
that when agents are self-interested, i.e., each agent is trained to
maximize its own reward, they only coordinate through the non-verbal
proposal channel, corroborating results of 
game-theoretical analyses \cite{Crawford:Sobel:1982}. On the other
hand, ``pro-social'' agents (that receive the
cumulative reward of both agents for the split they agreed upon) do
learn to meaningfully rely on the linguistic channel.

\citeA{Jaques:etal:2019} reported similar negative results for vanilla cheap talk among 
self-interested agents in the context of sequential social dilemmas \cite{Leibo:etal:2017}. 
Inspired by theories of the importance of social learning \cite{Herrmann:etal:2007}, the authors extended individual task rewards with an extra term capturing an agent \emph{social influence}, i.e., how effective the sender's active communication is on other agents. This is calculated as the impact that silencing the agent's communication channel has on the other agents' behaviour, and acts as intrinsic motivation for learning useful communication strategies. %
This inductive social bias results in agents with better
coordination skills, and consequently higher collective rewards.

\subsection{Machines Cooperating with Humans}
\label{sec:nll-learning}

One of the most ambitious goals of AI is to develop intelligent agents able to interact with humans.  Endowing agents with communication is an important milestone towards reaching this goal. Indeed, \citeA{Crandall:etal:2018} reported an experiment where humans had to coordinate with machines in repeated games \cite{Crandall:Goodrich:2010}. Importantly,  the machines were extended with scripted communication behaviour in natural language.  %
When cheap talk was not permitted, human-human and human-machine interactions rarely resulted in cooperation. Natural-language-based cheap talk, instead, increased cooperation and coordination in both cases.

The experimental setup of current multi-agent simulations, standard in machine learning, dictates stability of interlocutors between the training and testing phases, i.e., agents learn to communicate with a closed set of partners, and we then test their co-adaptation skills, or, to put it bluntly, how well they overfit their interlocutors. It is arguable whether advances in this setup will translate to genuine progress towards acquiring 
general communication skills transferable to other situations,
including interaction with different partners, such as
humans. \citeA{Carroll:etal:2019} studied humans cooperating with
machines trained via machine-machine (non-verbal) interaction, and
found that transfer from machine-machine to machine-human is not trivial. Agents co-adapt by establishing very idiosyncratic
conventions, since they possess different cognitive biases from
humans. We expect similar phenomena to also arise when cooperation
involves emergent communication protocols, which, as discussed above, often
develop counterintuitive properties.

Agent talk would be more easily generalizable, particularly in human-machine communication scenarios, if it were somehow aligned with natural language from the start. %
Since the dominating approach to natural language processing consists in passively extracting statistical generalizations from large amount of human-generated text \cite<e.g.,>{Radford:etal:2019}, thus guaranteeing alignment with the latter, some studies are starting to explore how to combine this approach with interactive multi-agent language learning~\cite{Lowe:etal:2020}. 
\citeA{Lazaridou:etal:2017}, inspired by analogous ideas in AI game
playing \cite{Silver:etal:2016}, explored a simple way to achieve this
combination, interleaving emergent communication and supervised
learning of names for a subset of the objects in their referential game (Fig.~\ref{fig:referential-game}). Under this mixed training
regime, the ``words'' used by the agents, even to refer to categories
for which the sender received no direct supervision, were generally
interpretable by humans. Interesting semantic shift phenomena also
emerged, such as the use of ``metonymic'' reference (using the word
for dolphin to refer to the sea). 

The main challenge when combining interactive multi-agent learning with natural language is language drift, i.e., the fact that pressures from the multi-agent tasks push protocols away from human language. 
 \citeA{Havrylov:Titov:2017} pre-trained a language model on English text corpora, and used its
statistics to constrain the emergent protocol. In practice, this meant
that the agents' utterances were fluent and grammatical, however there
was no constraint to align word meanings with English (i.e., the agent
word \emph{dogs} could refer to cats). 
To alleviate this nuisance, \citeA{Lee:etal:2019} explicitly enforced grounded alignment with natural
language by combining emergent communication and supervised
caption generation in a multi-task setup.  \citeA{Lu:etal:2020} take inspiration from the iterated learning paradigm in laboratory simulations of language evolution \cite{Kirby:Hurford:2002}. The first generation of learners starts with an agent that is pre-trained on task-specific natural language data using supervised learning and subsequently fine-tuned using rewards generated within a multi-agent framework. Each subsequent generation of learners is then pre-trained on samples of the language generated by the previous generation. %
~\citeA{Lazaridou:etal:2020} directly equip agents with a pre-trained general (i.e., not task-specific) image-conditioned language model, and use the rewards generated through multi-agent interaction to steer it towards the functional aspects of the particular task the agents are faced with (a vision-based referential game). Using pre-trained language models helps alleviate aspects of drift related to syntax and semantics. However, human evaluation shows that learning to use natural language within this multi-agent framework leads to pragmatic drift phenomena, where agents' and humans' contextual utterance interpretation might differ (e.g.,  \textit{Mike has a hat} is interpreted by agents as meaning \textit{Mike has a yellow hat} in a context where this inference is not valid). 



Aiming for realistic applications involving human-machine communication, such as natural language dialogue, future work should bridge the gap between the primitive communication
needs of agents in emergent language simulations, typically satisfied
by a code consisting of single words or short sentences, and the
grammatical nuance deep networks can exhibit when trained with large-scale language
modeling.  


\section{Concluding Remarks}
\label{sec:conclusion}

We surveyed the many active fronts of research in
multi-agent emergent language, empowered by advances in deep learning. Current simulations have become more realistic, running in 
setups which often include real-world images or grounded 3D environments,
giving rise to protocols with several intriguing properties. We conclude here by briefly listing a number of open questions and exciting directions for future work.

From a more theoretical perspective, we hope that more researchers from AI, linguistics and cognitive science will chip in 
with stronger hypotheses to shape experimental and analytical work, as
well as to provide insights into how to make computational models more
human-like.  Concepts playing an important role in the study of human communication and language, such as joint attention, theory
of mind and syntactic recursion are currently under-studied in the field of
multi-agent communication.  Hopefully, interdisciplinary insights will help modelers inject the right biases into agent
architectures and experimental setups. This will in turn result in simulations
that are not only more realistic in terms of investigating the roots of natural language, but also more
relevant to the kind of situations agents would encounter in actual human interaction.  Fortunately, toolkits are becoming available that facilitate
 entry into the area by scientists from other disciplines \cite<e.g.,>{Kharitonov:etal:2019}. 

While the studies presented here have already introduced a number of methods for
inspecting emergent protocols, important open issues remain with respect to how
to analyze emergent languages and whether it is possible to develop automated tools
that could speed up and generalize their analysis.

An even more serious issue that future research must address is that,
in the vast majority of current simulations, the
communication partners of agents during training and testing phases are the
same, i.e., we evaluate communication between agents that were trained
together. As such, it is not clear to what extent we are evaluating agents on
their ability to develop general communication skills, or simply their ability
to co-adapt to their learning environment and partners.

Perhaps the next big frontier with respect to
Artificial Intelligence lies in bringing the emergent language of agents to a
level of complexity and generality that will make them useful in applications.
However, just like human language did not probably emerge all at once,
we should outline precise desiderata for a minimally useful agent
\emph{proto-language}. At the same time, the huge progress currently being made
in corpus-based statistical natural language learning should be harnessed
to encourage the emergence of more interpretable and fluent protocols, thus
making multi-agent communication an integral part of human-centric AI.

\section*{Acknowledgements}

We would like to thank Joel Leibo, Phil Blunsom, Amanpreet Singh and Kyunghyun Cho for comments on earlier versions of this document and Chris Dyer, Rahma Chaabouni, Evgeny Kharitonov, Diane Bouchacourt and Emmanuel Dupoux for useful discussions. All images that were not created by us or reprinted by permission as indicated in the captions are in the public domain.

\bibliographystyle{theapa}
\bibliography{marco,angeliki}

\begin{thebibliography}{}

\bibitem[\protect\BCAY{Allwood}{Allwood}{1976}]{Allwood:1976}
Allwood, J. \BBOP1976\BBCP.
\newblock {\Bem Linguistic Communication as Action and Cooperation}.
\newblock Phd thesis, University of Goteborg.

\bibitem[\protect\BCAY{Anderson, He, Buehler, Teney, Johnson, Gould,\ \BBA\
  Zhang}{Anderson et~al.}{2018}]{Anderson:etal:2018}
Anderson, P., He, X., Buehler, C., Teney, D., Johnson, M., Gould, S., \BBA\
  Zhang, L. \BBOP2018\BBCP.
\newblock \BBOQ Bottom-up and top-down attention for image captioning and
  visual question answering\BBCQ\
\newblock In {\Bem Proceedings of CVPR}, \BPGS\ 6077--6086, Salt Lake City, UT.

\bibitem[\protect\BCAY{Andreas}{Andreas}{2019}]{Andreas:2019}
Andreas, J. \BBOP2019\BBCP.
\newblock \BBOQ Measuring compositionality in representation learning\BBCQ\
\newblock In {\Bem Proceedings of ICLR}, New Orleans, LA.
\newblock Published online:
  \url{https://openreview.net/group?id=ICLR.cc/2019/conference}.

\bibitem[\protect\BCAY{Austin}{Austin}{1962}]{Austin:1962}
Austin, J.~L. \BBOP1962\BBCP.
\newblock {\Bem How to do things with words}.
\newblock Harvard University Press, Cambridge, MA.

\bibitem[\protect\BCAY{Bakker, {Daval-Markussen}, Parkvall,\ \BBA\ Plag}{Bakker
  et~al.}{2011}]{Bakker:etal:2011}
Bakker, P., {Daval-Markussen}, A., Parkvall, M., \BBA\ Plag, I. \BBOP2011\BBCP.
\newblock \BBOQ Creoles are typologically distinct from non-creoles\BBCQ\
\newblock {\Bem Journal of Pidgin andCreole Languages}, {\Bem 26}, 5--42.

\bibitem[\protect\BCAY{Bard, Foerster, Chandar, Burch, Lanctot, Song,
  Parisotto, Dumoulin, Moitra, Hughes, et~al.}{Bard
  et~al.}{2020}]{Bard:etal:2020}
Bard, N., Foerster, J.~N., Chandar, S., Burch, N., Lanctot, M., Song, H.~F.,
  Parisotto, E., Dumoulin, V., Moitra, S., Hughes, E., et~al. \BBOP2020\BBCP.
\newblock \BBOQ The {Hanabi} challenge: A new frontier for {AI} research\BBCQ\
\newblock {\Bem Artificial Intelligence}, {\Bem 280}, 103216.

\bibitem[\protect\BCAY{Batali}{Batali}{1998}]{Batali:1998}
Batali, J. \BBOP1998\BBCP.
\newblock \BBOQ Computational simulations of the emergence of grammar\BBCQ\
\newblock In Hurford, J., {Studdert-Kennedy}, M., \BBA\ Knight, C.\BEDS, {\Bem
  Approaches to the Evolution of Language: Social and Cognitive Bases}, \BPGS\
  405--426. Cambridge University Press, Cambridge, UK.

\bibitem[\protect\BCAY{Bernardi, Boleda, Fern{\'a}ndez,\ \BBA\
  Paperno}{Bernardi et~al.}{2015}]{Bernardi:etal:2015}
Bernardi, R., Boleda, G., Fern{\'a}ndez, R., \BBA\ Paperno, D. \BBOP2015\BBCP.
\newblock \BBOQ Distributional semantics in use\BBCQ\
\newblock In {\Bem Proceedings of the EMNLP Workshop on Linking Computational
  Models of Lexical, Sentential and Discourse-level Semantics}, \BPGS\ 95--101,
  Lisbon, Portugal.

\bibitem[\protect\BCAY{Bickerton}{Bickerton}{2014}]{Bickerton:2014}
Bickerton, D. \BBOP2014\BBCP.
\newblock {\Bem More than Nature Needs: Language, Mind, and Evolution}.
\newblock Harvard University Press, Cambridge, MA.

\bibitem[\protect\BCAY{Bogin, Geva,\ \BBA\ Berant}{Bogin
  et~al.}{2018}]{Bogin:etal:2018}
Bogin, B., Geva, M., \BBA\ Berant, J. \BBOP2018\BBCP.
\newblock \BBOQ Emergence of communication in an interactive world with
  consistent speakers\BBCQ\
\newblock In {\Bem Proceedings of the NeurIPS Emergent Communication Workshop},
  Montreal, Canada.
\newblock Published online: \url{https://arxiv.org/abs/1809.00549}.

\bibitem[\protect\BCAY{Bouchacourt\ \BBA\ Baroni}{Bouchacourt\ \BBA\
  Baroni}{2018}]{Bouchacourt:Baroni:2018}
Bouchacourt, D.\BBACOMMA\  \BBA\ Baroni, M. \BBOP2018\BBCP.
\newblock \BBOQ How agents see things: On visual representations in an emergent
  language game\BBCQ\
\newblock In {\Bem Proceedings of EMNLP}, \BPGS\ 981--985, Brussels, Belgium.

\bibitem[\protect\BCAY{Bouchacourt\ \BBA\ Baroni}{Bouchacourt\ \BBA\
  Baroni}{2019}]{Bouchacourt:Baroni:2019}
Bouchacourt, D.\BBACOMMA\  \BBA\ Baroni, M. \BBOP2019\BBCP.
\newblock \BBOQ {Miss Tools and Mr Fruit}: {Emergent} communication in agents
  learning about object affordances\BBCQ\
\newblock In {\Bem Proceedings of ACL}, \BPGS\ 3909--3918, Florence, Italy.

\bibitem[\protect\BCAY{Bratman, Shvartsman, Lewis,\ \BBA\ Singh}{Bratman
  et~al.}{2010}]{Bratman:etal:2010}
Bratman, J., Shvartsman, M., Lewis, R., \BBA\ Singh, S. \BBOP2010\BBCP.
\newblock \BBOQ A new approach to exploring language emergence as boundedly
  optimal control in the face of environmental and cognitive constraints\BBCQ\
\newblock In {\Bem Proceedings of ICCM}, \BPGS\ 7--12, Philadelphia, PA.

\bibitem[\protect\BCAY{Brighton\ \BBA\ Kirby}{Brighton\ \BBA\
  Kirby}{2006}]{Brighton:2006}
Brighton, H.\BBACOMMA\  \BBA\ Kirby, S. \BBOP2006\BBCP.
\newblock \BBOQ Understanding linguistic evolution by visualizing the emergence
  of topographic mappings\BBCQ\
\newblock {\Bem Artificial life}, {\Bem 12\/}(2), 229--242.

\bibitem[\protect\BCAY{Cangelosi\ \BBA\ Parisi}{Cangelosi\ \BBA\
  Parisi}{2002}]{Cangelosi:Parisi:2002}
Cangelosi, A.\BBACOMMA\  \BBA\ Parisi, D.\BEDS. \BBOP2002\BBCP.
\newblock {\Bem Simulating the evolution of language}.
\newblock Springer, New York.

\bibitem[\protect\BCAY{Cao, Lazaridou, Lanctot, Leibo, Tuyls,\ \BBA\ Clark}{Cao
  et~al.}{2018}]{Cao:etal:2018}
Cao, K., Lazaridou, A., Lanctot, M., Leibo, J., Tuyls, K., \BBA\ Clark, S.
  \BBOP2018\BBCP.
\newblock \BBOQ Emergent communication through negotiation\BBCQ\
\newblock In {\Bem Proceedings of ICLR Conference Track}, Vancouver, Canada.
\newblock Published online:
  \url{https://openreview.net/group?id=ICLR.cc/2018/Conference}.

\bibitem[\protect\BCAY{Carroll, Shah, Ho, Griffiths, Seshia, Abbeel,\ \BBA\
  Dragan}{Carroll et~al.}{2019}]{Carroll:etal:2019}
Carroll, M., Shah, R., Ho, M.~K., Griffiths, T., Seshia, S., Abbeel, P., \BBA\
  Dragan, A. \BBOP2019\BBCP.
\newblock \BBOQ On the utility of learning about humans for human-ai
  coordination\BBCQ\
\newblock In {\Bem Advances in Neural Information Processing Systems}, \BPGS\
  5175--5186.

\bibitem[\protect\BCAY{Chaabouni, Kharitonov, Bouchacourt, Dupoux,\ \BBA\
  Baroni}{Chaabouni et~al.}{2020}]{Chaabouni:etal:2020}
Chaabouni, R., Kharitonov, E., Bouchacourt, D., Dupoux, E., \BBA\ Baroni, M.
  \BBOP2020\BBCP.
\newblock \BBOQ Compositionality and generalization in emergent languages\BBCQ\
\newblock In {\Bem Proceedings of ACL}, virtual conference.
\newblock {I}n press.

\bibitem[\protect\BCAY{Chaabouni, Kharitonov, Dupoux,\ \BBA\ Baroni}{Chaabouni
  et~al.}{2019}]{Chaabouni:etal:2019}
Chaabouni, R., Kharitonov, E., Dupoux, E., \BBA\ Baroni, M. \BBOP2019\BBCP.
\newblock \BBOQ Anti-efficient encoding in emergent communication\BBCQ\
\newblock In {\Bem Proceedings of NeurIPS}, Vancouver, Canada.
\newblock Published online:
  \url{https://papers.nips.cc/book/advances-in-neural-information-processing-systems-32-2019}.

\bibitem[\protect\BCAY{Choi, Lazaridou,\ \BBA\ {de Freitas}}{Choi
  et~al.}{2018}]{Choi:etal:2018}
Choi, E., Lazaridou, A., \BBA\ {de Freitas}, N. \BBOP2018\BBCP.
\newblock \BBOQ Compositional obverter communication learning from raw visual
  input\BBCQ\
\newblock In {\Bem Proceedings of ICLR Conference Track}, Vancouver, Canada.
\newblock Published online:
  \url{https://openreview.net/group?id=ICLR.cc/2018/Conference}.

\bibitem[\protect\BCAY{Christiansen\ \BBA\ Kirby}{Christiansen\ \BBA\
  Kirby}{2003}]{Christiansen:Kirby:2003}
Christiansen, M.\BBACOMMA\  \BBA\ Kirby, S.\BEDS. \BBOP2003\BBCP.
\newblock {\Bem Language Evolution}.
\newblock Oxford University Press, Oxford, UK.

\bibitem[\protect\BCAY{Clark}{Clark}{1996}]{Clark:1996}
Clark, H. \BBOP1996\BBCP.
\newblock {\Bem Using Language}.
\newblock Cambridge University Press, Cambridge, UK.

\bibitem[\protect\BCAY{Cortes, Martinez, Karatas,\ \BBA\ Bullo}{Cortes
  et~al.}{2004}]{Cortes:etal:2004}
Cortes, J., Martinez, S., Karatas, T., \BBA\ Bullo, F. \BBOP2004\BBCP.
\newblock \BBOQ Coverage control for mobile sensing networks\BBCQ\
\newblock {\Bem IEEE Transactions on robotics and Automation}, {\Bem 20\/}(2),
  243--255.

\bibitem[\protect\BCAY{Crandall\ \BBA\ Goodrich}{Crandall\ \BBA\
  Goodrich}{2011}]{Crandall:Goodrich:2010}
Crandall, J.\BBACOMMA\  \BBA\ Goodrich, M. \BBOP2011\BBCP.
\newblock \BBOQ Learning to compete, coordinate, and cooperate in repeated
  games using reinforcement learning\BBCQ\
\newblock {\Bem Machine Learning}, {\Bem 82}, 281--314.

\bibitem[\protect\BCAY{Crandall, Oudah, Ishowo-Oloko, Abdallah, Bonnefon,
  Cebrian, Shariff, Goodrich, Rahwan, et~al.}{Crandall
  et~al.}{2018}]{Crandall:etal:2018}
Crandall, J.~W., Oudah, M., Ishowo-Oloko, F., Abdallah, S., Bonnefon, J.-F.,
  Cebrian, M., Shariff, A., Goodrich, M.~A., Rahwan, I., et~al. \BBOP2018\BBCP.
\newblock \BBOQ Cooperating with machines\BBCQ\
\newblock {\Bem Nature communications}, 1--12.

\bibitem[\protect\BCAY{Crawford\ \BBA\ Sobel}{Crawford\ \BBA\
  Sobel}{1982}]{Crawford:Sobel:1982}
Crawford, V.~P.\BBACOMMA\  \BBA\ Sobel, J. \BBOP1982\BBCP.
\newblock \BBOQ Strategic information transmission\BBCQ\
\newblock {\Bem Econometrica: Journal of the Econometric Society}, 1431--1451.

\bibitem[\protect\BCAY{Das, Gervet, Romoff, Batra, Parikh, Rabbat,\ \BBA\
  Pineau}{Das et~al.}{2019}]{Das:etal:2019}
Das, A., Gervet, T., Romoff, J., Batra, D., Parikh, D., Rabbat, M., \BBA\
  Pineau, J. \BBOP2019\BBCP.
\newblock \BBOQ {T}ar{MAC}: Targeted multi-agent communication\BBCQ\
\newblock In {\Bem Proceedings of ICML}, \BPGS\ 1538--1546, Long Beach, CA.

\bibitem[\protect\BCAY{{David-Barrett}\ \BBA\ Dunbar}{{David-Barrett}\ \BBA\
  Dunbar}{2016}]{DavidBarrett:Dunbar:2016}
{David-Barrett}, T.\BBACOMMA\  \BBA\ Dunbar, R. \BBOP2016\BBCP.
\newblock \BBOQ Language as a coordination tool evolves slowly\BBCQ\
\newblock {\Bem Royal Society Open Science}, {\Bem 3\/}(12), 160259.

\bibitem[\protect\BCAY{DeVault, Mell,\ \BBA\ Gratch}{DeVault
  et~al.}{2015}]{DeVault:etal:2015}
DeVault, D., Mell, J., \BBA\ Gratch, J. \BBOP2015\BBCP.
\newblock \BBOQ Toward natural turn-taking in a virtual human negotiation
  agent\BBCQ\
\newblock In {\Bem 2015 AAAI Spring Symposium Series}.

\bibitem[\protect\BCAY{Devlin, Chang, Lee,\ \BBA\ Toutanova}{Devlin
  et~al.}{2019}]{Devlin:etal:2019}
Devlin, J., Chang, M.-W., Lee, K., \BBA\ Toutanova, K. \BBOP2019\BBCP.
\newblock \BBOQ {BERT}: {Pre-training} of deep bidirectional transformers for
  language understanding\BBCQ\
\newblock In {\Bem Proceedings of NAACL}, \BPGS\ 4171--4186, Minneapolis, MN.

\bibitem[\protect\BCAY{Eccles, Bachrach, Lever, Lazaridou,\ \BBA\
  Graepel}{Eccles et~al.}{2019}]{Eccles:etal:2019}
Eccles, T., Bachrach, Y., Lever, G., Lazaridou, A., \BBA\ Graepel, T.
  \BBOP2019\BBCP.
\newblock \BBOQ Biases for emergent communication in multi-agent reinforcement
  learning\BBCQ\
\newblock In {\Bem Advances in Neural Information Processing Systems}, \BPGS\
  13111--13121.

\bibitem[\protect\BCAY{Edunov, Ott, Auli,\ \BBA\ Grangier}{Edunov
  et~al.}{2018}]{Edunov:etal:2018}
Edunov, S., Ott, M., Auli, M., \BBA\ Grangier, D. \BBOP2018\BBCP.
\newblock \BBOQ Understanding back-translation at scale\BBCQ\
\newblock In {\Bem Proceedings of EMNLP}, \BPGS\ 489--500, Brussels, Belgium.

\bibitem[\protect\BCAY{Elman}{Elman}{1990}]{Elman:1990}
Elman, J. \BBOP1990\BBCP.
\newblock \BBOQ Finding structure in time\BBCQ\
\newblock {\Bem Cognitive Science}, {\Bem 14}, 179--211.

\bibitem[\protect\BCAY{Evtimova, Drozdov, Kiela,\ \BBA\ Cho}{Evtimova
  et~al.}{2018}]{Evtimova:etal:2018}
Evtimova, K., Drozdov, A., Kiela, D., \BBA\ Cho, K. \BBOP2018\BBCP.
\newblock \BBOQ Emergent communication in a multi-modal, multi-step referential
  game\BBCQ\
\newblock In {\Bem Proceedings of ICLR Conference Track}, Vancouver, Canada.
\newblock Published online:
  \url{https://openreview.net/group?id=ICLR.cc/2018/Conference}.

\bibitem[\protect\BCAY{Farrell\ \BBA\ Rabin}{Farrell\ \BBA\
  Rabin}{1996}]{Farrell:Rabin:1996}
Farrell, J.\BBACOMMA\  \BBA\ Rabin, M. \BBOP1996\BBCP.
\newblock \BBOQ Cheap talk\BBCQ\
\newblock {\Bem Journal of Economic Perspectives}, {\Bem 10\/}(3), 103--118.

\bibitem[\protect\BCAY{{Ferrer i Cancho}, Hern{\'a}ndez-Fern{\'a}ndez, Lusseau,
  Agoramoorthy, Hsu,\ \BBA\ Semple}{{Ferrer i Cancho}
  et~al.}{2013}]{FerrerICancho:etal:2013}
{Ferrer i Cancho}, R., Hern{\'a}ndez-Fern{\'a}ndez, A., Lusseau, D.,
  Agoramoorthy, G., Hsu, M., \BBA\ Semple, S. \BBOP2013\BBCP.
\newblock \BBOQ Compression as a universal principle of animal behavior\BBCQ\
\newblock {\Bem Cognitive Science}, {\Bem 37\/}(8), 1565--1578.

\bibitem[\protect\BCAY{Foerster, Assael, de~Freitas,\ \BBA\ Whiteson}{Foerster
  et~al.}{2016}]{Foerster:etal:2016}
Foerster, J., Assael, I.~A., de~Freitas, N., \BBA\ Whiteson, S. \BBOP2016\BBCP.
\newblock \BBOQ Learning to communicate with deep multi-agent reinforcement
  learning\BBCQ\
\newblock In {\Bem Proceedings of NIPS}, \BPGS\ 2137--2145, Barcelona, Spain.

\bibitem[\protect\BCAY{Foerster, Song, Hughes, Burch, Dunning, Whiteson,
  Botvinick,\ \BBA\ Bowling}{Foerster et~al.}{2019}]{Foerster:etal:2019}
Foerster, J.~N., Song, F., Hughes, E., Burch, N., Dunning, I., Whiteson, S.,
  Botvinick, M., \BBA\ Bowling, M. \BBOP2019\BBCP.
\newblock \BBOQ Bayesian action decoder for deep multi-agent reinforcement
  learning\BBCQ\
\newblock {\Bem International Conference on Machine Learning}.

\bibitem[\protect\BCAY{Gao, Galley,\ \BBA\ Li}{Gao
  et~al.}{2019}]{Gao:etal:2019}
Gao, J., Galley, M., \BBA\ Li, L. \BBOP2019\BBCP.
\newblock {\Bem Neural Approaches to Conversational {AI}: Question Answering,
  Task-Oriented Dialogues and Social Chatbots}.
\newblock Now Publishers, Norwell, MA.

\bibitem[\protect\BCAY{Gibson, Piantadosi, Dautriche, Mahowald, Bergen,\ \BBA\
  Levy}{Gibson et~al.}{2019}]{Gibson:etal:2019}
Gibson, E., Piantadosi, R. F.~S., Dautriche, I., Mahowald, K., Bergen, L.,
  \BBA\ Levy, R. \BBOP2019\BBCP.
\newblock \BBOQ How efficiency shapes human language\BBCQ\
\newblock {\Bem Trends in Cognitive Science}.
\newblock {I}n press.

\bibitem[\protect\BCAY{Ginzburg\ \BBA\ Poesio}{Ginzburg\ \BBA\
  Poesio}{2016}]{Ginzburg:Poesio:2016}
Ginzburg, J.\BBACOMMA\  \BBA\ Poesio, M. \BBOP2016\BBCP.
\newblock \BBOQ Grammar is a system that characterizes talk in
  interaction\BBCQ\
\newblock {\Bem Frontiers in Psychology}, {\Bem 7\/}(1938), 1--22.

\bibitem[\protect\BCAY{Goodman\ \BBA\ Frank}{Goodman\ \BBA\
  Frank}{2016}]{Goodman:Frank:2016}
Goodman, N.~D.\BBACOMMA\  \BBA\ Frank, M.~C. \BBOP2016\BBCP.
\newblock \BBOQ Pragmatic language interpretation as probabilistic
  inference\BBCQ\
\newblock {\Bem Trends in cognitive sciences}, {\Bem 20\/}(11), 818--829.

\bibitem[\protect\BCAY{Graesser, Cho,\ \BBA\ Kiela}{Graesser
  et~al.}{2019}]{Graesser:etal:2019}
Graesser, L., Cho, K., \BBA\ Kiela, D. \BBOP2019\BBCP.
\newblock \BBOQ Emergent linguistic phenomena in multi-agent communication
  games\BBCQ\
\newblock In {\Bem Proceedings of EMNLP}, \BPGS\ 3700--3710, Hong Kong, China.

\bibitem[\protect\BCAY{Grice}{Grice}{1975}]{Grice:1975}
Grice, H.~P. \BBOP1975\BBCP.
\newblock \BBOQ Logic and conversation\BBCQ\
\newblock In {\Bem Speech acts}, \BPGS\ 41--58. Brill.

\bibitem[\protect\BCAY{G{\"u}th, Schmittberger,\ \BBA\ Schwarze}{G{\"u}th
  et~al.}{1982}]{Guth:etal:1982}
G{\"u}th, W., Schmittberger, R., \BBA\ Schwarze, B. \BBOP1982\BBCP.
\newblock \BBOQ An experimental analysis of ultimatum bargaining\BBCQ\
\newblock {\Bem Journal of economic behavior \& organization}, {\Bem 3\/}(4),
  367--388.

\bibitem[\protect\BCAY{Havrylov\ \BBA\ Titov}{Havrylov\ \BBA\
  Titov}{2017}]{Havrylov:Titov:2017}
Havrylov, S.\BBACOMMA\  \BBA\ Titov, I. \BBOP2017\BBCP.
\newblock \BBOQ Emergence of language with multi-agent games: Learning to
  communicate with sequences of symbols\BBCQ\
\newblock In {\Bem Proceedings of NIPS}, \BPGS\ 2149--2159, Long Beach, CA.

\bibitem[\protect\BCAY{Herrmann, Call, Hern{\'a}ndez-Lloreda, Hare,\ \BBA\
  Tomasello}{Herrmann et~al.}{2007}]{Herrmann:etal:2007}
Herrmann, E., Call, J., Hern{\'a}ndez-Lloreda, M.~V., Hare, B., \BBA\
  Tomasello, M. \BBOP2007\BBCP.
\newblock \BBOQ Humans have evolved specialized skills of social cognition: The
  cultural intelligence hypothesis\BBCQ\
\newblock {\Bem science}, {\Bem 317\/}(5843), 1360--1366.

\bibitem[\protect\BCAY{Hochreiter\ \BBA\ Schmidhuber}{Hochreiter\ \BBA\
  Schmidhuber}{1997}]{Hochreiter:Schmidhuber:1997}
Hochreiter, S.\BBACOMMA\  \BBA\ Schmidhuber, J. \BBOP1997\BBCP.
\newblock \BBOQ Long short-term memory\BBCQ\
\newblock {\Bem Neural Computation}, {\Bem 9\/}(8), 1735--1780.

\bibitem[\protect\BCAY{Hockett}{Hockett}{1960}]{Hockett:1960}
Hockett, C. \BBOP1960\BBCP.
\newblock \BBOQ The origin of speech\BBCQ\
\newblock {\Bem Scientific American}, {\Bem 203}, 88--111.

\bibitem[\protect\BCAY{Hurford}{Hurford}{2014}]{Hurford:2014}
Hurford, J. \BBOP2014\BBCP.
\newblock {\Bem The Origins of Language}.
\newblock Oxford University Press, Oxford, UK.

\bibitem[\protect\BCAY{Jang, Gu,\ \BBA\ Poole}{Jang
  et~al.}{2017}]{Jang:etal:2017}
Jang, E., Gu, S., \BBA\ Poole, B. \BBOP2017\BBCP.
\newblock \BBOQ Categorical reparameterization with {Gumbel-Softmax}\BBCQ\
\newblock In {\Bem Proceedings of ICLR Conference Track}, Toulon, France.
\newblock Published online:
  \url{https://openreview.net/group?id=ICLR.cc/2017/conference}.

\bibitem[\protect\BCAY{Jaques, Lazaridou, Hughes, Gulcehre, Ortega, Strouse,
  Leibo,\ \BBA\ De~Freitas}{Jaques et~al.}{2019}]{Jaques:etal:2019}
Jaques, N., Lazaridou, A., Hughes, E., Gulcehre, C., Ortega, P., Strouse, D.,
  Leibo, J., \BBA\ De~Freitas, N. \BBOP2019\BBCP.
\newblock \BBOQ Social influence as intrinsic motivation for multi-agent deep
  reinforcement learning\BBCQ\
\newblock In {\Bem Proceedings of ICML}, \BPGS\ 3040--3049, Long Beach, CA.

\bibitem[\protect\BCAY{Jorge, K{\aa}geb{\"a}ck,\ \BBA\ Gustavsson}{Jorge
  et~al.}{2016}]{Jorge:etal:2016}
Jorge, E., K{\aa}geb{\"a}ck, M., \BBA\ Gustavsson, E. \BBOP2016\BBCP.
\newblock \BBOQ Learning to play {Guess Who?}~and inventing a grounded language
  as a consequence\BBCQ\
\newblock In {\Bem Proceedings of the NIPS Deep Reinforcement Learning
  Workshop}, Barcelona, Spain.
\newblock Published online:
  \url{https://sites.google.com/site/deeprlnips2016/}.

\bibitem[\protect\BCAY{Kasai, Tenmoto,\ \BBA\ Kamiya}{Kasai
  et~al.}{2008}]{Kasai:etal:2008}
Kasai, T., Tenmoto, H., \BBA\ Kamiya, A. \BBOP2008\BBCP.
\newblock \BBOQ Learning of communication codes in multi-agent reinforcement
  learning problem\BBCQ\
\newblock In {\Bem 2008 IEEE Conference on Soft Computing in Industrial
  Applications}, \BPGS\ 1--6. IEEE.

\bibitem[\protect\BCAY{Kharitonov, Chaabouni, Bouchacourt,\ \BBA\
  Baroni}{Kharitonov et~al.}{2019}]{Kharitonov:etal:2019}
Kharitonov, E., Chaabouni, R., Bouchacourt, D., \BBA\ Baroni, M.
  \BBOP2019\BBCP.
\newblock \BBOQ {EGG}: a toolkit for research on emergence of lan{g}uage in
  games\BBCQ\
\newblock In {\Bem Proceedings of EMNLP (System Demonstrations)}, \BPGS\
  55--60, Hong Kong, China.

\bibitem[\protect\BCAY{Kharitonov, Chaabouni, Bouchacourt,\ \BBA\
  Baroni}{Kharitonov et~al.}{2020}]{Kharitonov:etal:2020}
Kharitonov, E., Chaabouni, R., Bouchacourt, D., \BBA\ Baroni, M.
  \BBOP2020\BBCP.
\newblock \BBOQ Entropy minimization in emergent languages\BBCQ\
\newblock In {\Bem Proceedings of ICML}, virtual conference.
\newblock {I}n press.

\bibitem[\protect\BCAY{Kim, Moon, Hostallero, Kang, Lee, Son,\ \BBA\ Yi}{Kim
  et~al.}{2019}]{Kim:etal:2019}
Kim, D., Moon, S., Hostallero, D., Kang, W.~J., Lee, T., Son, K., \BBA\ Yi, Y.
  \BBOP2019\BBCP.
\newblock \BBOQ Learning to schedule communication in multi-agent reinforcement
  learning\BBCQ\
\newblock In {\Bem Proceedings of ICLR}, New Orleans, LA.
\newblock Published online:
  \url{https://openreview.net/group?id=ICLR.cc/2019/conference}.

\bibitem[\protect\BCAY{Kirby, Griffiths,\ \BBA\ Smith}{Kirby
  et~al.}{2014}]{Kirby:etal:2014}
Kirby, S., Griffiths, T., \BBA\ Smith, K. \BBOP2014\BBCP.
\newblock \BBOQ Iterated learning and the evolution of language\BBCQ\
\newblock {\Bem Current Opinion in Neurobiology}, {\Bem 28}, 108--114.

\bibitem[\protect\BCAY{Kirby\ \BBA\ Hurford}{Kirby\ \BBA\
  Hurford}{2002}]{Kirby:Hurford:2002}
Kirby, S.\BBACOMMA\  \BBA\ Hurford, J. \BBOP2002\BBCP.
\newblock \BBOQ The emergence of linguistic structure: An overview of the
  iterated learning model\BBCQ\
\newblock In Cangelosi, A.\BBACOMMA\  \BBA\ Parisi, D.\BEDS, {\Bem Simulating
  the evolution of language}. Springer, New York.

\bibitem[\protect\BCAY{Kottur, Moura, Lee,\ \BBA\ Batra}{Kottur
  et~al.}{2017}]{Kottur:etal:2017}
Kottur, S., Moura, J., Lee, S., \BBA\ Batra, D. \BBOP2017\BBCP.
\newblock \BBOQ Natural language does not emerge `naturally' in multi-agent
  dialog\BBCQ\
\newblock In {\Bem Proceedings of EMNLP}, \BPGS\ 2962--2967, Copenhagen,
  Denmark.

\bibitem[\protect\BCAY{Krizhevsky, Sutskever,\ \BBA\ Hinton}{Krizhevsky
  et~al.}{2017}]{Krizhevsky:etal:2017}
Krizhevsky, A., Sutskever, I., \BBA\ Hinton, G. \BBOP2017\BBCP.
\newblock \BBOQ {ImageNet} classification with deep convolutional neural
  networks\BBCQ\
\newblock {\Bem Communications of the ACM}, {\Bem 60\/}(6), 84--90.

\bibitem[\protect\BCAY{Lazaridou, Hermann, Tuyls,\ \BBA\ Clark}{Lazaridou
  et~al.}{2018}]{Lazaridou:etal:2018}
Lazaridou, A., Hermann, K., Tuyls, K., \BBA\ Clark, S. \BBOP2018\BBCP.
\newblock \BBOQ Emergence of linguistic communication from referential games
  with symbolic and pixel input\BBCQ\
\newblock In {\Bem Proceedings of ICLR Conference Track}, Vancouver, Canada.
\newblock Published online:
  \url{https://openreview.net/group?id=ICLR.cc/2018/Conference}.

\bibitem[\protect\BCAY{Lazaridou, Peysakhovich,\ \BBA\ Baroni}{Lazaridou
  et~al.}{2017}]{Lazaridou:etal:2017}
Lazaridou, A., Peysakhovich, A., \BBA\ Baroni, M. \BBOP2017\BBCP.
\newblock \BBOQ Multi-agent cooperation and the emergence of (natural)
  language\BBCQ\
\newblock In {\Bem Proceedings of ICLR Conference Track}, Toulon, France.
\newblock Published online:
  \url{https://openreview.net/group?id=ICLR.cc/2017/conference}.

\bibitem[\protect\BCAY{Lazaridou, Potapenko,\ \BBA\ Tieleman}{Lazaridou
  et~al.}{2020}]{Lazaridou:etal:2020}
Lazaridou, A., Potapenko, A., \BBA\ Tieleman, O. \BBOP2020\BBCP.
\newblock \BBOQ Multi-agent communication meets natural language: Synergies
  between functional and structural language learning\BBCQ\
\newblock {\Bem Association of Computational Linguistics}.

\bibitem[\protect\BCAY{LeCun, Bengio,\ \BBA\ Hinton}{LeCun
  et~al.}{2015}]{LeCun:etal:2015}
LeCun, Y., Bengio, Y., \BBA\ Hinton, G. \BBOP2015\BBCP.
\newblock \BBOQ Deep learning\BBCQ\
\newblock {\Bem Nature}, {\Bem 521}, 436--444.

\bibitem[\protect\BCAY{LeCun, Bottou, Bengio,\ \BBA\ Haffner}{LeCun
  et~al.}{1998}]{LeCun:etal:1998}
LeCun, Y., Bottou, L., Bengio, Y., \BBA\ Haffner, P. \BBOP1998\BBCP.
\newblock \BBOQ Gradient-based learning applied to document recognition\BBCQ\
\newblock {\Bem Proceedings of the IEEE}, {\Bem 86\/}(11), 2278--2324.

\bibitem[\protect\BCAY{Lee, Cho,\ \BBA\ Kiela}{Lee
  et~al.}{2019}]{Lee:etal:2019}
Lee, J., Cho, K., \BBA\ Kiela, D. \BBOP2019\BBCP.
\newblock \BBOQ Countering language drift via visual grounding\BBCQ\
\newblock In {\Bem Proceedings of the 2019 Conference on Empirical Methods in
  Natural Language Processing and the 9th International Joint Conference on
  Natural Language Processing (EMNLP-IJCNLP)}, \BPGS\ 4376--4386.

\bibitem[\protect\BCAY{Leibo, Zambaldi, Lanctot, Marecki,\ \BBA\ Graepel}{Leibo
  et~al.}{2017}]{Leibo:etal:2017}
Leibo, J.~Z., Zambaldi, V., Lanctot, M., Marecki, J., \BBA\ Graepel, T.
  \BBOP2017\BBCP.
\newblock \BBOQ Multi-agent reinforcement learning in sequential social
  dilemmas\BBCQ\
\newblock {\Bem Proceedings of the 16th International Conference on Autonomous
  Agents and Multiagent Systems}.

\bibitem[\protect\BCAY{Lewis}{Lewis}{1969}]{Lewis:1969}
Lewis, D. \BBOP1969\BBCP.
\newblock {\Bem Convention}.
\newblock Harvard University Press, Cambridge, MA.

\bibitem[\protect\BCAY{Li\ \BBA\ Bowling}{Li\ \BBA\
  Bowling}{2019}]{Li:Bowling:2019}
Li, F.\BBACOMMA\  \BBA\ Bowling, M. \BBOP2019\BBCP.
\newblock \BBOQ Ease-of-teaching and language structure from emergent
  communication\BBCQ\
\newblock In {\Bem Proceedings of NeurIPS}, Vancouver, Canada.
\newblock Published online:
  \url{https://papers.nips.cc/book/advances-in-neural-information-processing-systems-32-2019}.

\bibitem[\protect\BCAY{Linell}{Linell}{2009}]{Linell:2009}
Linell, P. \BBOP2009\BBCP.
\newblock {\Bem Rethinking Language, Mind, and World Dialogically:
  Interactional and Contextual Theories of Human Sense-making}.
\newblock Information Age Publishers, Charlotte, NC.

\bibitem[\protect\BCAY{Lowe, Foerster, Boureau, Pineau,\ \BBA\ Dauphin}{Lowe
  et~al.}{2019}]{Lowe:etal:2019}
Lowe, R., Foerster, J., Boureau, Y., Pineau, J., \BBA\ Dauphin, Y.
  \BBOP2019\BBCP.
\newblock \BBOQ On the pitfalls of measuring emergent communication\BBCQ\
\newblock In {\Bem Proceedings of AAMAS}, \BPGS\ 693--701, Montreal, Canada.

\bibitem[\protect\BCAY{Lowe, Gupta, Foerster, Kiela,\ \BBA\ Pineau}{Lowe
  et~al.}{2020}]{Lowe:etal:2020}
Lowe, R., Gupta, A., Foerster, J., Kiela, D., \BBA\ Pineau, J. \BBOP2020\BBCP.
\newblock \BBOQ On the interaction between supervision and self-play in
  emergent communication\BBCQ\
\newblock {\Bem International Conference on Learning Representation}.

\bibitem[\protect\BCAY{Lowe, Wu, Tamar, Harb, Abbeel,\ \BBA\ Mordatch}{Lowe
  et~al.}{2017}]{Lowe:etal:2017}
Lowe, R., Wu, Y., Tamar, A., Harb, J., Abbeel, P., \BBA\ Mordatch, I.
  \BBOP2017\BBCP.
\newblock \BBOQ Multi-agent actor-critic for mixed cooperative-competitive
  environments\BBCQ\
\newblock In {\Bem Advances in neural information processing systems}, \BPGS\
  6379--6390.

\bibitem[\protect\BCAY{Lu, Singhal, Strub, Pietquin,\ \BBA\ Courville}{Lu
  et~al.}{2020}]{Lu:etal:2020}
Lu, Y., Singhal, S., Strub, F., Pietquin, O., \BBA\ Courville, A.
  \BBOP2020\BBCP.
\newblock \BBOQ Countering language drift with seeded iterated learning\BBCQ\
\newblock {\Bem International Conference on Machine Learning}.

\bibitem[\protect\BCAY{Lupyan\ \BBA\ Bergen}{Lupyan\ \BBA\
  Bergen}{2016}]{Lupyan:Bergen:2016}
Lupyan, G.\BBACOMMA\  \BBA\ Bergen, B. \BBOP2016\BBCP.
\newblock \BBOQ How language programs the mind\BBCQ\
\newblock {\Bem Topics in Cognitive Science}, {\Bem 8}, 408--424.

\bibitem[\protect\BCAY{Lux\ \BBA\ Marchesi}{Lux\ \BBA\
  Marchesi}{1999}]{Lux:Marchesi:1999}
Lux, T.\BBACOMMA\  \BBA\ Marchesi, M. \BBOP1999\BBCP.
\newblock \BBOQ Scaling and criticality in a stochastic multi-agent model of a
  financial market\BBCQ\
\newblock {\Bem Nature}, {\Bem 397\/}(6719), 498--500.

\bibitem[\protect\BCAY{Maddison, Mnih,\ \BBA\ Teh}{Maddison
  et~al.}{2017}]{Maddison:etal:2017}
Maddison, C., Mnih, A., \BBA\ Teh, Y. \BBOP2017\BBCP.
\newblock \BBOQ The concrete distribution: {A} continuous relaxation of
  discrete random variables\BBCQ\
\newblock In {\Bem Proceedings of ICLR Conference Track}, Toulon, France.
\newblock Published online:
  \url{https://openreview.net/group?id=ICLR.cc/2017/conference}.

\bibitem[\protect\BCAY{Mikolov, Joulin,\ \BBA\ Baroni}{Mikolov
  et~al.}{2016}]{Mikolov:etal:2016}
Mikolov, T., Joulin, A., \BBA\ Baroni, M. \BBOP2016\BBCP.
\newblock \BBOQ A roadmpap towards machine intelligence\BBCQ\
\newblock In {\Bem Proceedings of CICLing}, \BPGS\ 29--61.

\bibitem[\protect\BCAY{Mnih, Kavukcuoglu, Silver, Rusu, Veness, Bellemare,
  Graves, Riedmiller, Fidjeland, Ostrovski, Petersen, Beattie, Sadik,
  Antonoglou, King, Kumaran, Wierstra, Legg,\ \BBA\ Hassabis}{Mnih
  et~al.}{2015}]{Mnih:etal:2015}
Mnih, V., Kavukcuoglu, K., Silver, D., Rusu, A., Veness, J., Bellemare, M.,
  Graves, A., Riedmiller, M., Fidjeland, A., Ostrovski, G., Petersen, S.,
  Beattie, C., Sadik, A., Antonoglou, I., King, H., Kumaran, D., Wierstra, D.,
  Legg, S., \BBA\ Hassabis, D. \BBOP2015\BBCP.
\newblock \BBOQ Human-level control through deep reinforcement learning\BBCQ\
\newblock {\Bem Nature}, {\Bem 518}, 529--533.

\bibitem[\protect\BCAY{Mordatch\ \BBA\ Abbeel}{Mordatch\ \BBA\
  Abbeel}{2018}]{Mordatch:Abbeel:2018}
Mordatch, I.\BBACOMMA\  \BBA\ Abbeel, P. \BBOP2018\BBCP.
\newblock \BBOQ Emergence of grounded compositional language in multi-agent
  populations\BBCQ\
\newblock In {\Bem Proceedings of AAAI}, \BPGS\ 1495--1502, New Orleans, LA.

\bibitem[\protect\BCAY{{Myers-Scotton}}{{Myers-Scotton}}{2002}]{MyersScotton:2002}
{Myers-Scotton}, C. \BBOP2002\BBCP.
\newblock {\Bem Contact Linguistics: {Bilingual} Encounters and Grammatical
  Outcomes}.
\newblock Oxford University Press, Oxford, UK.

\bibitem[\protect\BCAY{Panait\ \BBA\ Luke}{Panait\ \BBA\
  Luke}{2005}]{Panait:Luke:2005}
Panait, L.\BBACOMMA\  \BBA\ Luke, S. \BBOP2005\BBCP.
\newblock \BBOQ Cooperative multi-agent learning: The state of the art\BBCQ\
\newblock {\Bem Autonomous agents and multi-agent systems}, {\Bem 11\/}(3),
  387--434.

\bibitem[\protect\BCAY{Pickering\ \BBA\ Garrod}{Pickering\ \BBA\
  Garrod}{2004}]{Pickering:Garrod:2004}
Pickering, M.\BBACOMMA\  \BBA\ Garrod, S. \BBOP2004\BBCP.
\newblock \BBOQ Toward a mechanistic psychology of dialogue\BBCQ\
\newblock {\Bem Behavorial and Brain Sciences}, {\Bem 27\/}(2), 169--190.

\bibitem[\protect\BCAY{Radford, Wu, Child, Luan, Amodei,\ \BBA\
  Sutskever}{Radford et~al.}{2019}]{Radford:etal:2019}
Radford, A., Wu, J., Child, R., Luan, D., Amodei, D., \BBA\ Sutskever, I.
  \BBOP2019\BBCP.
\newblock \BBOQ Language models are unsupervised multitask learners\BBCQ\
\newblock
  \url{https://d4mucfpksywv.cloudfront.net/better-language-models/language-models.pdf}.

\bibitem[\protect\BCAY{Raviv, Meyer,\ \BBA\ Lev-Ari}{Raviv
  et~al.}{2019a}]{Raviv:etal:2019a}
Raviv, L., Meyer, A., \BBA\ Lev-Ari, S. \BBOP2019a\BBCP.
\newblock \BBOQ Compositional structure can emerge without generational
  transmission\BBCQ\
\newblock {\Bem Cognition}, {\Bem 182}, 151--164.

\bibitem[\protect\BCAY{Raviv, Meyer,\ \BBA\ Lev-Ari}{Raviv
  et~al.}{2019b}]{Raviv:etal:2019b}
Raviv, L., Meyer, A., \BBA\ Lev-Ari, S. \BBOP2019b\BBCP.
\newblock \BBOQ Larger communities create more systematic languages\BBCQ\
\newblock {\Bem Proceedings of the Royal Society B}, {\Bem 286\/}(1907),
  20191262.

\bibitem[\protect\BCAY{Ren, Guo, Havrylov, Cohen,\ \BBA\ Kirby}{Ren
  et~al.}{2019}]{Ren:etal:2019}
Ren, Y., Guo, S., Havrylov, S., Cohen, S., \BBA\ Kirby, S. \BBOP2019\BBCP.
\newblock \BBOQ Enhance the compositionality of emergent language by iterated
  learning\BBCQ\
\newblock In {\Bem Proceedings of the NeurIPS Emergent Communication Workshop},
  Vancouver, Canada.
\newblock Published online:
  \url{https://sites.google.com/view/emecom2019/accepted-papers}.

\bibitem[\protect\BCAY{Resnick, Gupta, Foerster, Dai,\ \BBA\ Cho}{Resnick
  et~al.}{2020}]{Resnick:etal:2020}
Resnick, C., Gupta, A., Foerster, J., Dai, A., \BBA\ Cho, K. \BBOP2020\BBCP.
\newblock \BBOQ Capacity, bandwidth, and compositionality in emergent language
  learning\BBCQ\
\newblock In {\Bem Proceedings of AAMAS}, Auckland, New Zealand.
\newblock {I}n press.

\bibitem[\protect\BCAY{Russakovsky, Deng, Su, Krause, Satheesh, Ma, Huang,
  Karpathy, Khosla, Bernstein, Berg,\ \BBA\ Fei{-}Fei}{Russakovsky
  et~al.}{2015}]{Russakovsky:etal:2014}
Russakovsky, O., Deng, J., Su, H., Krause, J., Satheesh, S., Ma, S., Huang, Z.,
  Karpathy, A., Khosla, A., Bernstein, M., Berg, A., \BBA\ Fei{-}Fei, L.
  \BBOP2015\BBCP.
\newblock \BBOQ {ImageNet Large Scale Visual Recognition} challenge\BBCQ\
\newblock {\Bem International Journal of Computer Vision}, {\Bem 115\/}(3),
  211--252.

\bibitem[\protect\BCAY{Searle}{Searle}{1969}]{Searle:1969}
Searle, J. \BBOP1969\BBCP.
\newblock {\Bem Speech Acts: An Essay in the Philosophy of Language}.
\newblock Cambridge University Press, Cambridge, UK.

\bibitem[\protect\BCAY{Serban, Lowe, Charlin,\ \BBA\ Pineau}{Serban
  et~al.}{2016}]{Serban:etal:2016}
Serban, I., Lowe, R., Charlin, L., \BBA\ Pineau, J. \BBOP2016\BBCP.
\newblock \BBOQ Generative deep neural networks for dialogue: {A} short
  review\BBCQ\
\newblock In {\Bem Proceedings of the NIPS Learning Methods for Dialogue
  Workshop}, Barcelona, Spain.
\newblock Published online:
  \url{http://letsdiscussnips2016.weebly.com/schedule.html}.

\bibitem[\protect\BCAY{Silver, Huang, Maddison, Guez, Sifre, {van den
  Driessche}, Schrittwieser, Antonoglou, Panneershelvam, Lanctot, Dieleman,
  Grewe, Nham, Kalchbrenner, Sutskever, Lillicrap, Leach, Kavukcuoglu,
  Graepel,\ \BBA\ Hassabis}{Silver et~al.}{2016}]{Silver:etal:2016}
Silver, D., Huang, A., Maddison, C., Guez, A., Sifre, L., {van den Driessche},
  G., Schrittwieser, J., Antonoglou, I., Panneershelvam, V., Lanctot, M.,
  Dieleman, S., Grewe, D., Nham, J., Kalchbrenner, N., Sutskever, I.,
  Lillicrap, T., Leach, M., Kavukcuoglu, K., Graepel, T., \BBA\ Hassabis, D.
  \BBOP2016\BBCP.
\newblock \BBOQ Mastering the game of {Go} with deep neural networks and tree
  search\BBCQ\
\newblock {\Bem Nature}, {\Bem 529}, 484--503.

\bibitem[\protect\BCAY{Singh, Jain,\ \BBA\ Sukhbaatar}{Singh
  et~al.}{2019}]{Singh:etal:2019}
Singh, A., Jain, T., \BBA\ Sukhbaatar, S. \BBOP2019\BBCP.
\newblock \BBOQ Learning when to communicate at scale in multiagent cooperative
  and competitive tasks\BBCQ\
\newblock In {\Bem Proceedings of ICLR}, New Orleans, LA.
\newblock Published online:
  \url{https://openreview.net/group?id=ICLR.cc/2019/conference}.

\bibitem[\protect\BCAY{Skyrms}{Skyrms}{2010}]{Skyrms:2010}
Skyrms, B. \BBOP2010\BBCP.
\newblock {\Bem Signals: Evolution, learning, and information}.
\newblock Oxford University Press, Oxford, UK.

\bibitem[\protect\BCAY{Steels}{Steels}{2003}]{Steels:2003b}
Steels, L. \BBOP2003\BBCP.
\newblock \BBOQ Evolving grounded communication for robots\BBCQ\
\newblock {\Bem Trends in Cognitive Sciences}, {\Bem 7\/}(7), 308--312.

\bibitem[\protect\BCAY{Steels}{Steels}{2012}]{Steels:2012}
Steels, L.\BED. \BBOP2012\BBCP.
\newblock {\Bem Experiments in Cultural Language Evolution}.
\newblock John Benjamins, Amsterdam, the Netherlands.

\bibitem[\protect\BCAY{Strauss, Grzybek,\ \BBA\ Altmann}{Strauss
  et~al.}{2007}]{Strauss:etal:2007}
Strauss, U., Grzybek, P., \BBA\ Altmann, G. \BBOP2007\BBCP.
\newblock \BBOQ Word length and word frequency\BBCQ\
\newblock In Grzybek, P.\BED, {\Bem Contributions to the Science of Text and
  Language}, \BPGS\ 277--294. Springer, Dordrecht, the Netherlands.

\bibitem[\protect\BCAY{Sukhbaatar, Szlam,\ \BBA\ Fergus}{Sukhbaatar
  et~al.}{2016}]{Sukhbaatar:etal:2016}
Sukhbaatar, S., Szlam, A., \BBA\ Fergus, R. \BBOP2016\BBCP.
\newblock \BBOQ Learning multiagent communication with backpropagation\BBCQ\
\newblock In {\Bem Proceedings of NIPS}, \BPGS\ 2244--2252, Barcelona, Spain.

\bibitem[\protect\BCAY{Suter, Miladinovic, Sch{\"o}lkopf,\ \BBA\ Bauer}{Suter
  et~al.}{2019}]{Suter:etal:2019}
Suter, R., Miladinovic, D., Sch{\"o}lkopf, B., \BBA\ Bauer, S. \BBOP2019\BBCP.
\newblock \BBOQ Robustly disentangled causal mechanisms: {V}alidating deep
  representations for interventional robustness\BBCQ\
\newblock In {\Bem Proceedings of ICML}, \BPGS\ 6056--6065, Long Beach, CA.

\bibitem[\protect\BCAY{Sutton\ \BBA\ Barto}{Sutton\ \BBA\
  Barto}{1998}]{Sutton:Barto:1998}
Sutton, R.\BBACOMMA\  \BBA\ Barto, A. \BBOP1998\BBCP.
\newblock {\Bem Reinforcement Learning: An Introduction}.
\newblock MIT Press, Cambridge, MA.

\bibitem[\protect\BCAY{Tan}{Tan}{1993}]{Tan:1993}
Tan, M. \BBOP1993\BBCP.
\newblock \BBOQ Multi-agent reinforcement learning: independent versus
  cooperative agents\BBCQ\
\newblock In {\Bem Proceedings of the Tenth International Conference on
  International Conference on Machine Learning}, \BPGS\ 330--337. Morgan
  Kaufmann Publishers Inc.

\bibitem[\protect\BCAY{Tieleman, Lazaridou, Mourad, Blundell,\ \BBA\
  Precup}{Tieleman et~al.}{2019}]{Tieleman:etal:2019}
Tieleman, O., Lazaridou, A., Mourad, S., Blundell, C., \BBA\ Precup, D.
  \BBOP2019\BBCP.
\newblock \BBOQ Shaping representations through communication: community size
  effect in artificial learning systems\BBCQ\
\newblock {\Bem NeurIPS workshop on Visually Grounded Interaction and
  Language}.

\bibitem[\protect\BCAY{Tomasello}{Tomasello}{2010}]{Tomasello:2010}
Tomasello, M. \BBOP2010\BBCP.
\newblock {\Bem Origins of Human Communication}.
\newblock MIT Press, Cambridge, MA.

\bibitem[\protect\BCAY{Townsend, Engesser, Stoll, Zuberb\"{u}hler,\ \BBA\
  Bickel}{Townsend et~al.}{2018}]{Townsend:etal:2018}
Townsend, S., Engesser, S., Stoll, S., Zuberb\"{u}hler, K., \BBA\ Bickel, B.
  \BBOP2018\BBCP.
\newblock \BBOQ Compositionality in animals and humans\BBCQ\
\newblock {\Bem PLOS Biology}, {\Bem 16\/}(8), 1--7.

\bibitem[\protect\BCAY{Vaswani, Shazeer, Parmar, Uszkoreit, Jones, Gomez,
  Kaiser,\ \BBA\ Polosukhin}{Vaswani et~al.}{2017}]{Vaswani:etal:2017}
Vaswani, A., Shazeer, N., Parmar, N., Uszkoreit, J., Jones, L., Gomez, A.,
  Kaiser, L., \BBA\ Polosukhin, I. \BBOP2017\BBCP.
\newblock \BBOQ Attention is all you need\BBCQ\
\newblock In {\Bem Proceedings of NIPS}, \BPGS\ 5998--6008, Long Beach, CA.

\bibitem[\protect\BCAY{Von~Frisch}{Von~Frisch}{1967}]{VonFrisch:1967}
Von~Frisch, K. \BBOP1967\BBCP.
\newblock
\newblock \BBOQ The dance language and orientation of bees.\BBCQ.

\bibitem[\protect\BCAY{Wagner, Reggia, Uriagereka,\ \BBA\ Wilkinson}{Wagner
  et~al.}{2003}]{Wagner:etal:2003}
Wagner, K., Reggia, J., Uriagereka, J., \BBA\ Wilkinson, G. \BBOP2003\BBCP.
\newblock \BBOQ Progress in the simulation of emergent communication and
  language\BBCQ\
\newblock {\Bem Adaptive Behavior}, {\Bem 11\/}(1), 37--69.

\bibitem[\protect\BCAY{Williams}{Williams}{1992}]{Williams:1992}
Williams, R. \BBOP1992\BBCP.
\newblock \BBOQ Simple statistical gradient-following algorithms for
  connectionist reinforcement learning\BBCQ\
\newblock {\Bem Machine learning}, {\Bem 8\/}(3-4), 229--256.

\bibitem[\protect\BCAY{Wittgenstein}{Wittgenstein}{1953}]{Wittgenstein:1953}
Wittgenstein, L. \BBOP1953\BBCP.
\newblock {\Bem Philosophical Investigations}.
\newblock Blackwell, Oxford, UK.
\newblock Translated by G.E.M. Anscombe.

\bibitem[\protect\BCAY{Zhou, Palangi, Zhang, Hu, Corso,\ \BBA\ Gao}{Zhou
  et~al.}{2020}]{Zhou:etal:2020}
Zhou, L., Palangi, H., Zhang, L., Hu, H., Corso, J., \BBA\ Gao, J.
  \BBOP2020\BBCP.
\newblock \BBOQ Unified vision-language pre-training for image captioning and
  {VQA}\BBCQ\
\newblock In {\Bem Proceedings of AAAI}, New York, NY.
\newblock {I}n press.

\bibitem[\protect\BCAY{Zipf}{Zipf}{1949}]{Zipf:1949}
Zipf, G. \BBOP1949\BBCP.
\newblock {\Bem Human Behavior and the Principle of Least Effort}.
\newblock Addison-Wesley, Boston, MA.

\end{thebibliography}

\end{document}